%% file: drift.tex
\documentclass[11pt]{amsart}

\usepackage{tikz}
\usepackage{tkz-berge}
\usepackage{algpseudocode}
\usepackage{algorithm}
\usepackage{todonotes}
\usepackage{float}

\usetikzlibrary{patterns}
\usetikzlibrary{positioning}
\usetikzlibrary{patterns}
\usetikzlibrary{matrix}
\usetikzlibrary{shapes}
\usetikzlibrary{decorations.pathreplacing}
\usetikzlibrary{graphs,graphs.standard}
\usetikzlibrary{fadings}
\usepackage{tikz-3dplot}
\usepackage{hyperref}
\usepackage[labelfont=bf]{caption}
\GraphInit[vstyle=Classic]
\SetGraphUnit{1}
\SetVertexMath
\tikzset{EdgeStyle/.style = {bend right, font=\scriptsize, ->}} 
\tikzset{VertexStyle/.style = {shape = circle,fill = black,minimum size = 1pt, node font=\scriptsize}}

\newtheorem{thm}{Theorem}[section]

\theoremstyle{definition}
\newtheorem{example}[thm]{Example}
\newtheorem{remark}[thm]{Remark}
\newtheorem{defn}[thm]{Definition}

\numberwithin{equation}{section}

\newcommand{\qref}[1]{(\ref{#1})}

\newcommand{\Dlargemsingdrift}{\texttt{dataset-1}}
\newcommand{\Dlargemuldrift}{\texttt{dataset-2}}
\newcommand{\Dextrememuldrift}{\texttt{dataset-3}}

\usepackage{amsaddr}
\makeatletter
\renewcommand{\email}[2][]{%
  \ifx\emails\@empty\relax\else{\g@addto@macro\emails{,\space}}\fi%
  \@ifnotempty{#1}{\g@addto@macro\emails{\textrm{(#1)}\space}}%
  \g@addto@macro\emails{#2}%
}
\makeatother

\makeatletter
\def\algbackskip{\hskip-\ALG@thistlm}
\makeatother

\DeclareMathOperator*{\argmin}{arg\,min}
\newcommand\cD{{\mathcal D}}
\newcommand\cT{{\mathcal T}}
\def\eps{{\epsilon}}
\newcommand{\ring}[1]{\ensuremath{\mathbb{#1}}}
\newcommand\NN{\ring{N}}
\newcommand\RR{\ring{R}}

\title{A method to benchmark high-dimensional process drift detection}

\author{Edgar Wolf$^{\ast}$ \and Tobias Windisch}
\email{$\{$edgar.wolf,tobias.windisch$\}$@hs-kempten.de}
\email{$^{\ast}$\textnormal{Corresponding author}}

\address{University of Applied Sciences Kempten, Germany}

\begin{document}

\maketitle

\begin{abstract}
    Process curves are multivariate finite time series data coming
    from manufacturing processes. This paper studies machine learning
    that detect drifts in process curve datasets. A theoretic
    framework to synthetically generate process curves in a controlled
    way is introduced in order to benchmark machine learning
    algorithms for process drift detection. An evaluation score,
    called the temporal area under the curve, is introduced, which
    allows to quantify how well machine learning models unveil curves
    belonging to drift segments.  Finally, a benchmark study comparing
    popular machine learning approaches on synthetic data generated
    with the introduced framework is presented that shows that
    existing algorithms often struggle with datasets containing
    multiple drift segments. 
\end{abstract}

\section{Introduction}\label{s:intro}

\begin{figure}[htbp]
    \scalebox{0.8}{
    \input{./ts_drifts.tex}  
}
\caption{Overview of different kinds of time series data from
manufacturing processes and drifts within.}\label{f:different_drifts}\end{figure}
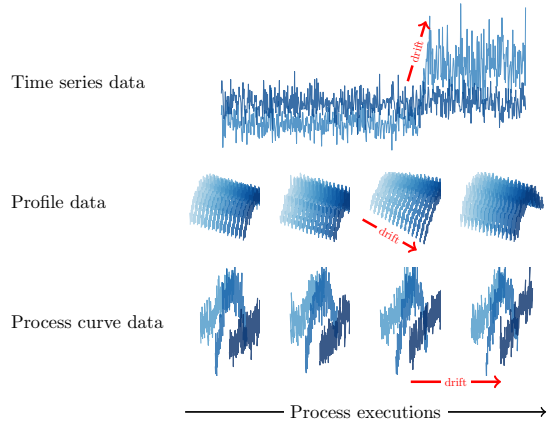

Manufacturing lines typically consist of \emph{processes} arranged
sequentially, each using
techniques like casting, forming, or joining to shape components to
their final specifications. Advanced sensor technology enables precise
monitoring of key performance indicators, like force, pressure, or
temperature, over time. IoT-enabled systems now commonly store the
data obtained, called \emph{process curves}, facilitating analysis across both single components
and entire production sequences~\cite{manufacturing_i4.0}.
Issues like anomalous batches, tool wear, or miscalibrations can
degrade performance, often subtly, by causing gradual shifts in
process curves. Thus, detecting \emph{process drifts} is key to keep unplanned
downtimes and scrap parts at bay.
In high-volume production, this is particularly
challenging due to the rapid data generation and complexity of
multi-variable curves and hence these settings have been an ideal
application for machine learning methods~\cite{tool_wear_monitoring_with_neural_networks,
    process_curves_casting_with_nn,ml_in_production, inline_drift_detection_laser_melting,
bayesian_ae_drift_detection_industrial_env,process_curves_feature_extraction,screw_fastening}.
Although process curves are multivariate time-series, process drift
detection should not be confused with drift detection in time
series~\cite{benchmark_time_series_drift} or drifts in profile
data~\cite{profile_data_characterization} (see also
Figure~\ref{f:different_drifts})). Typically, statistical
drift detection methods from time series analysis are not direct
applicable, not alone because process curves are high-dimensional
objects, but also because of high autocorrelation among their sample
axis. As a consequence,
deep learning techniques, most prominently dimensionality reduction
methods like \emph{autoencoders}~\cite{autoencoder,vae,iVAE, ica_aux_vars,
kimVariationalAutoencoderSemiconductor2023}, have become
increasingly popular~\cite{process_curves_in_manufacturing} as they
allow to first learn a
low-dimensional representation of the high-dimensional input
and then
analyse the learned latent variables with classic statistical tools, like sliding
Kolmogorov-Smirnov tests~\cite{kswin},
Hellinger-distance based techniques~\cite{h3dm}, or by facilitating the Maximum Mean
Discrepancy~\cite{mmd}. 
For many machine learning applications relevant for manufacturing,
established ways and datasets to benchmark the performance of algorithms exist,
like for \emph{causal discovery} in quality data~\cite{causalassembly},
\emph{anomaly detection} in images from optical inspections~\cite{mvtec},
or \emph{reinforcement
learning} in continuous control tasks~\cite{benchmark_rl}. However, for process drift detection,
such a framework is yet missing to the best of our knowledge. This may
be due to the following two
reasons: The lack of both, publicly available datasets and a suitable
evaluation metric. Beside
a few publicly released
datasets~\cite{nasa_milling_data, bosch_cnc_milling,
mauthe2024overview}, 
most of the existing work does not release any data to the
public, making it impossible to test other detectors on the same
dataset. Often, this is due to privacy issues and fear of leaking
information to competitors. 
In addition, as datasets for process drifts are inherently non
identically and identically distributed (iid), any sort of test and
train splits introduced significant biases making evaluation of
algorithms hard if only one variant of a dataset is available
Finally,
process drift detection is by definition an unsupervised learning task, but to
benchmark detectors, a ground truth is required, labeling precisely
when a drift starts and when it ends.
This seamlessly leads to the second challenge, namely the missing
evaluation metric. As in any machine learning task, the metric depends
on the precise application and a trustworthy ground truth.  A commonly
used metric used in research and practice to measure the statistical performance of a binary
classifier, often independent of the application, is the \emph{area
under the ROC curve}~\cite{auc} - short AUC. However, the usage of the AUC is
typically only applicable in settings where data is assumed to be iid,
unlike in drift detection.
In our work, we want to exactly address these issues by introducing 
a benchmarking framework for researchers, allowing them to
reliably validate their process drift detection algorithms. At a
high level, our main contributions are:
\begin{itemize}
    \item{We present a simple, yet flexible and effective theoretic framework to
generate synthetic process curve datasets including drifts with a validated
ground truth (Section~\ref{s:setup} and
Section~\ref{s:data-generation}) which also allows feeding of
curves from real processes.}
\item{We introduce an evaluation metric called \emph{temporal area
under the curve} (TAUC) in Section~\ref{s:tauc}, which aims to take the temporal context of a
detection into account.}
\item{We conduct a short benchmark study
in Section~\ref{s:experiments} as a proof of concept for the
effectiveness of both, our TAUC metric and our proposed data
generation method to measure the predictive power of
drift detectors.}
\end{itemize}
Our work is based on preliminary results
of the first author~\cite{ma_edgar}. In this work, we provide additionally insights into the introduced
metric, introduce a variant called \emph{soft TAUC} and compare it in
depth with existing metrics. Moreover, we substantially generalize the data
synthetization framework, for instance by allowing higher-order
derivatives, and we generate more
sophisticated datasets for the benchmark study. We also release the
code that helps to generate
process curves to benchmark drift detectors, which is freely
available under~\url{https://github.com/edgarWolf/driftbench}. 
Its optimization back-end is implemented in \texttt{Jax}~\cite{jax}
allowing a fast GPU-based generation of process curves.

\section{Statistical framework to model process drifts}\label{s:setup}

In this section, we formalize what we consider as process
curves and drifts within. Generally speaking, process curve datasets
are datasets consisting of finitely many
multivariate time series each having finitely many steps. 
\begin{figure}[htbp]
    \centering
    \scalebox{0.8}{
    \begin{tikzpicture}
    \node[] at (0,0) {\includegraphics[width=0.35\textwidth,trim={1cm 0cm 1cm 1cm}, clip]{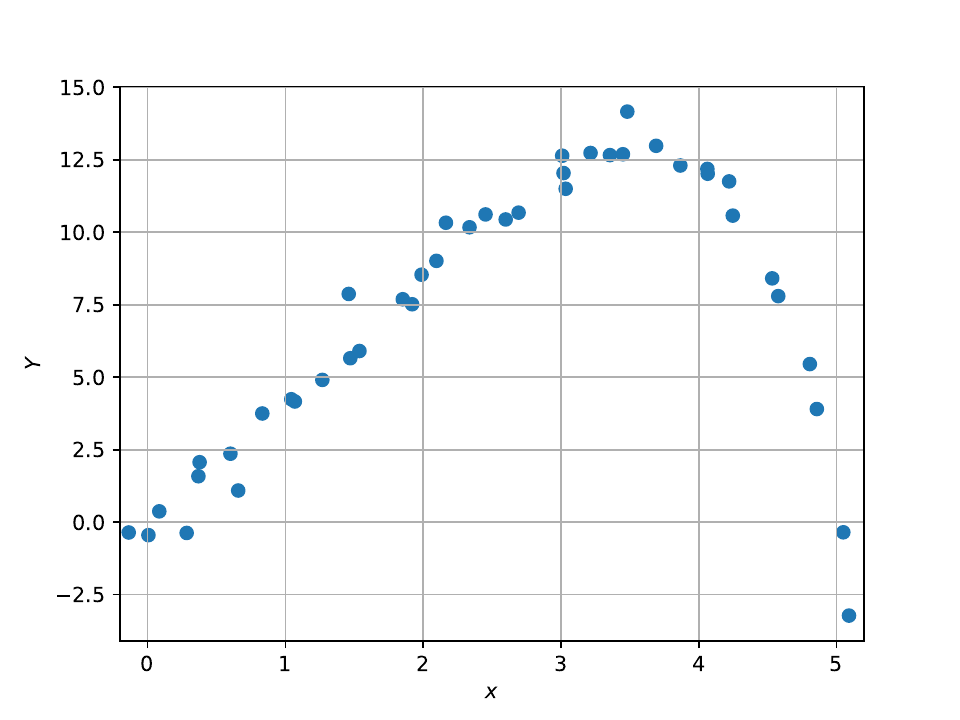}};
    \node[] at (6,0) {\includegraphics[width=0.45\textwidth,trim={1cm 1cm 1cm 1cm}, clip]{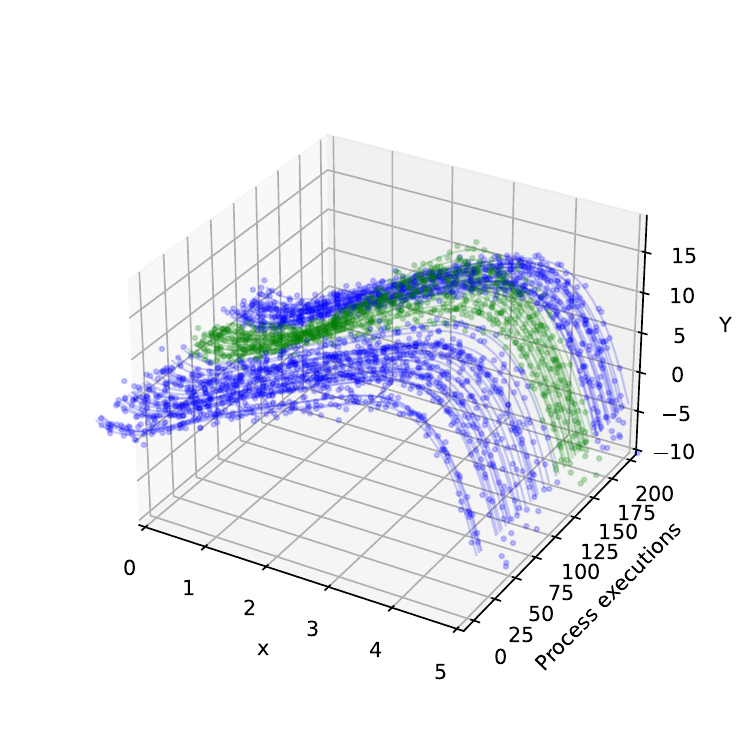}};
    \end{tikzpicture}
}
\caption{Samples from a process curve (left) as well as a sequence
of curve samples (right).}\label{f:drifts}\end{figure}

We formally model a process curve as a finite time-series $(Y(x))_{x\in I}$ with
$Y(x)\in\RR^c$, $I\subset\RR$ a finite set, and where $Y:\RR\to\RR^c$
represent physical properties of the process to be measured and $x$
an independent variable, often the time. In staking processes, for
instance, $Y$
is the
measured force and
$x$ the walked path of the press (compare
also~\cite[Figure~9]{causalassembly}). Another example are pneumatic
test stations, where $Y$
might be a pressure measured over time $x$. In bolt fastening
processes, $Y$ represents the torque measured over the angle
$x$~\cite{screw_fastening}. We call the number of variables $c\in\NN$
in the curve the \emph{dimension} of the process curve and write $[T]$
for the set $\{1,\ldots, T\}$ and often refer to it as \emph{temporal
axis}.

Whenever a manufacturing process finishes its work on a component, a
process curve is yielded. Thus, when the same process is executed on multiple
times sequentially, a long sequence
$C_1,C_2,\ldots,C_t,\ldots, C_T$ with $T\in\NN$ of process curves is obtained where each
$C_t$ arises under slightly different physical conditions $Y_1,\ldots, Y_T$ , i.e.,
$C_t=(Y_t(x+\eps_x)+\eps_y)_{x\in I_t}$, where $\eps_x$ and $\eps_y$
represents measurement noise or inaccuracies. In theory, also the sets $I_t$ can vary
for each $t\in[T]$, for instance due to different offsets. 
Wearout or tool degradation affects the process curves gradually and
to model their
deformation along the execution axis,
we assume that there exists functions
$f:\RR^k\times\RR\to\RR^c$ and
$w:[T]\to\RR^k$ such that for all $t\in[T]$ and $x\in
I_t$:
\begin{equation}\label{equ:data-gen}
f(w(t), x)=Y_t(x)
\end{equation}
where the function $f$ is a proxy for the physics underneath the
process. 
The vector $w(t)\in\RR^k$ represents environmental properties of the $t$-th
execution, and some of its coordinates correspond to component
properties, some to properties of the machine. Without restricting generality and to keep notation simple, we will
assume for the remainder that $c=1$, as the multivariate case is a
straight-forward application of our approach by modeling each variable
in $Y$ individually (see also Remark~\ref{r:multivariate-data}).

\begin{figure}[htbp]
    \centering
    \begin{tikzpicture}
        \node[] at (0,0) {\includegraphics[width=0.35\textwidth]{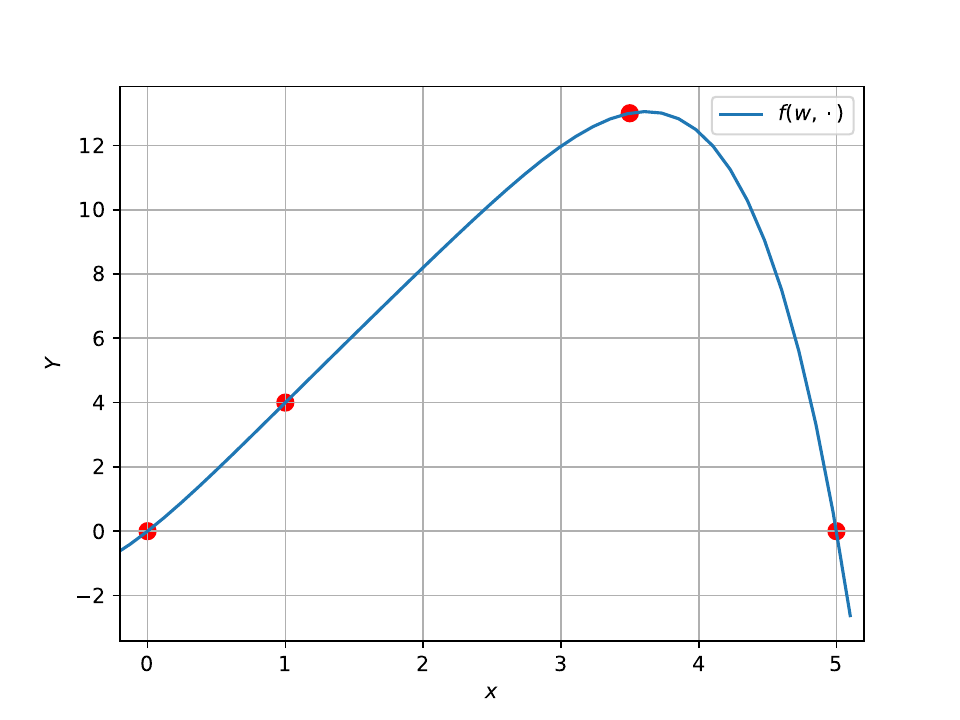}};
        \node[] at (7,0) {\includegraphics[width=0.35\textwidth]{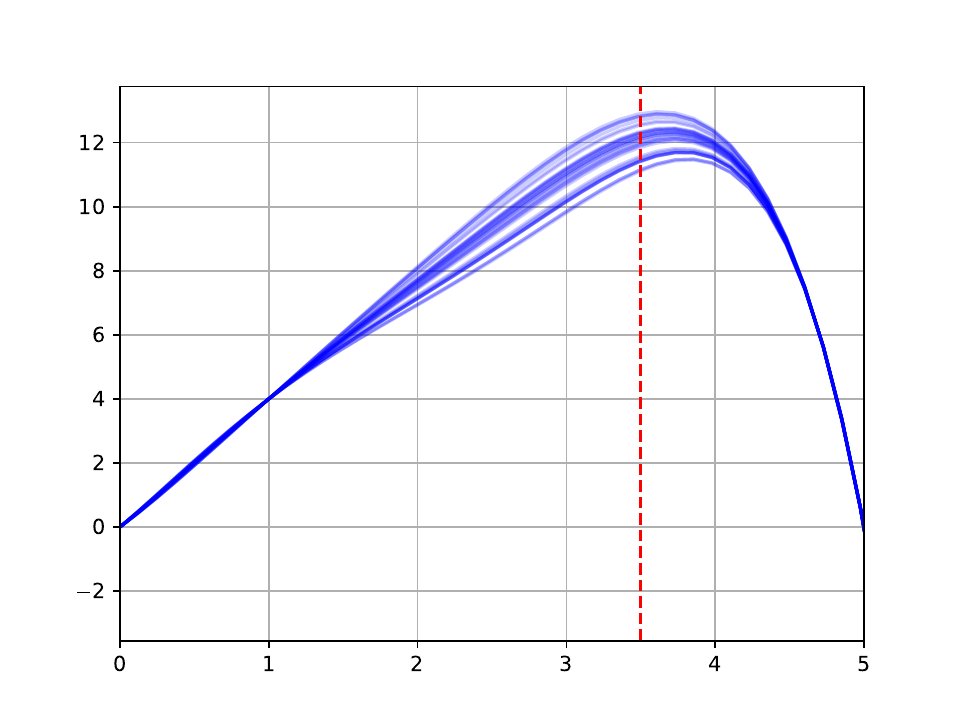}};
        
    \end{tikzpicture}
\caption{Visualization of the data synthetization given a function
$f(w, x)=\sum_{i=0}^5w_i\cdot x^i$. Left figure shows $f(w,\cdot)$
solved for concrete $x^i,y^i$ (red points). Right figure shows
sequence $f(w_1, \cdot),\ldots, f(w_{100},\cdot)$ where gaussian noise
was added on one coordinate in $y^1(t)$ (corresponding coordinate
$x^1(t)$ is marked with a dashed line).}\label{f:data_synthetization}
\end{figure}

Assuming only component variance and 
no tool degradation, we could
assume that $w(t)$ is sampled in each process execution from a fixed but unknown
distribution on $\RR^k$, like $w(t)\sim\mathcal{N}_{\mu,\sigma}$
with fixed $\mu\in\RR^k$ and $\sigma\in\RR^{k\times k}$ for all
$t\in[T]$. 
As mentioned, tool degradation, in contrast, affects the process from execution to
execution, i.e., the parameters of the distribution shift over time
leading to a deformation of the observed process curve. 
Such process drifts should not be confused with concept drifts, where
the goal is typically to analyse the declining performance of a
trained machine learning model when new data starts to differ from the
train data~\cite{cdd_typicality_eccentricity}. 
Moreover, detecting drifts in process curves is different to detecting
drift in \emph{profile data}~\cite{profile_data_characterization}, where one typically is interested in drifts among the curves yielded
by a single execution, not in drifts over multiple executions.
A similar application is the identification of drifts within profile data, where typically
one process execution
yields a sequence of process curves of fixed size, like in spectroscopy
when one curve is some intensity over time which is measured for
different wavelengths~\cite{profile_data}.
One way to model process drifts is to model the evolution of the
latent parameters $w(t)$, like using a dynamical system.
For instance, in control theory~\cite{controltheory}, $w(t)$ is considered as latent state of a system
which evolves over the executions $t$ and one observes a multivariate
output $Y(t)\in\RR^{|I_t|}$ with $Y(t)=Y_t(I_t)$.
Introducing a control vector $u(t)\in\RR^p$, $w(t)$
can be considered as state variable $w(t)$ of the system that
evolves over time and is influenced by a control vector $u(t)\in\RR^p$
such that $\partial_t w(t)=h(w(t), u(t),t)$ and $Y(t)=f(w(t))$ holds
for all $t\in\mathbb{N}$. Here, however, one has to precisely model
how the state $w$ changes over executions and how it is affected by
interventions $u$ and has to solve challenging non-linear differential
equations. However, as we will argue, the degradation of the curve can be
described directly in curve space in many scenarios. 
Thus, we directly model the transformation of the process curves in
curve space by letting certain \emph{support points} of the curve
move in a controlled way:
\begin{defn}[Support points]\label{def:support-point}
    Let $f:\RR^k\times\RR\to\RR$ be an $i$-times differentiable function, $i\in\NN$, and
    $\partial_x^i f$ be the $i$-th derivative of $f$ according to the
    second argument. Let $x,y\in\RR^n$,
    then 
    $(x,y)$ is a \emph{support point of $i$-th order} for $f$ at $w\in\RR^k$ if
    $\partial^{i}_x f(w, x_j)=y_j$ for all $j\in[n]$.
\end{defn}
Support points can be considered as points surpassed by
the graph of $f(w,\cdot):\RR\to\RR$ (see visualization on the
left in Figure~\ref{f:data_synthetization}). Typically, such support points are physically motivated and if
latent properties of the process 
change, certain support points change their position in curve space.
For instance, in a staking process, the position $x(t)$ and value
$y(t)$ of the
maximal force, i.e., where the first derivative is zero, starts
shifting (see Figure~\ref{f:drifts}). That is, we can describe this
behavior by modelling the support points $(x(t), y(t))$ and $(x(t), 0)$
of first and second order respectively, i.e.
$f(w(t), x(t))=y(t)$ and $\partial_x^1 f(w(t), x(t))=0$. We formalize in
Section~\ref{s:data-generation} how we can use this to generate
process curves and drifts synthetically.

\section{Data generation}\label{s:data-generation}

Let $f:\RR^k\times\RR\to\RR$ be as in Section~\ref{s:setup} a proxy for the physical
relations for given manufacturing process. In this section, we build our
synthetization framwork upon the
setup introduced in Section~\ref{s:setup}. Here, we neither focus on how $w(t)$
behaves in latent space, nor on how $f$ is formulated exactly. 
Instead of modeling the evolution of $w(t)$ with a dynamic system, our
idea is
to model the behavior of support points over process executions in curve space and
to seek for
parameters $w(t)$ using non-linear optimization satisfying the support
point conditions from Definition~\ref{def:support-point}. 
For the remainder of this section, we explain how $w(t)$ can be
computed given the support points.
Thus, assume we have for each process execution $t\in[T]$ 
support points $(x^1(t),y^1(t)),\ldots (x^l(t), y^l(t))$ with
with $x^i(t),y^i(t)\in\mathbb{R}^{n_i}$, that is,
\begin{equation}\label{equ:support-point-condition}
    \partial_x^i f(w(t), x^i_j(t)) = y^i_j(t) \quad\forall j\in [n_i].
\end{equation}

Instead of modelling $w(t)$ explicitly, we \emph{compute} $w(t)$
implicitly such
that~\qref{equ:support-point-condition} is satisfied. For instance,
if $f$ is $l+2$-times differentiable in its second argument and if $\partial_w^2 \partial_x^i f$ exists, we can
solve~\qref{equ:support-point-condition} individually
for all $t\in[T]$ using second-order quasi-Newton
methods~\cite[Chapter~3]{quasi_newton} for the objective function
\begin{equation}\label{equ:optimization-problem}
    w(t)=\argmin_{w\in\RR^k}\sum_{i=1}^l\sum_{j=1}^{n_i}D_i\cdot
    \left(\partial^i_x f(w, x^i_j(t))- y^i_j(t)\right)^2 
\end{equation}
where $D_1,\ldots,D_l$ are constants to account for the different
value ranges of the functions~$\partial^i_x f$. 
By solving
Thus, solving~\qref{equ:optimization-problem} for each $t\in[T]$,
we obtain a sequence $w(1),\ldots, w(T)\in\RR^k$ and
consequently, we get a sequence of functions $f(w(1),
\cdot),\ldots, f(w(t), \cdot)$. Now, these functions can be evaluated
on arbitrarily sets $I_t\subset\mathbb{R}$ whose point not necessarily need
to be equidistant. Setting $C_t=f(w(t), I_t)+\eps_y\in\RR^{|I_t|}$, we finally obtain a
sequence of process curves $C_1,\ldots, C_T$. A compact overview of
the data generation method is shown in Algorithm~\ref{a:data-gen}.
\begin{algorithm}
    \caption{Generation of process curves}
    \label{a:data-gen}
     \textbf{Input: }$f:\RR^k\times\RR\to\RR$,
     $x^i(1), y^i(1),\ldots x^i(T),y^i(T)\in\RR^{n_i}$,
     $i\in[l]$, $\overline{x}\in\RR$, $\Delta x\in\RR_{>0}$, $m\in\NN$\hfill\textcolor{white}{.}\\
     \textbf{Output: } Process curves $C_1,\ldots,
     C_T$\hfill\textcolor{white}{.}
    \begin{algorithmic}[1]
    \For{$t \in [T]$}
    \State Compute solution $w(t)$ for \qref{equ:optimization-problem}
    using support points $(x^1(t),y^1(t)),\ldots,(x^l(t),x^l(t))$
    \State $I_t\gets \{\overline{x}+j\cdot\Delta x+\eps_x: j\in[m]\}$
    \State $C_t\gets f(w(t), I_t)+\eps_y$
    \EndFor
    \State \Return $C_1,\ldots, C_T$
    \end{algorithmic}
\end{algorithm}

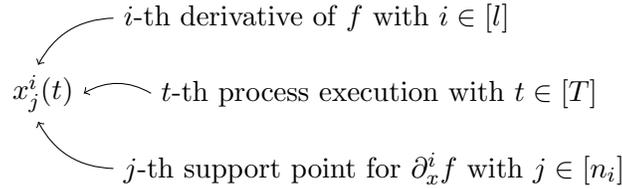
\begin{figure}[htbp]
    \centering
\begin{tikzpicture}
    \node[] (x) at (0,0) {~~$x_j^i(t)$};
    \node[anchor=west] (j) at (1, -1) {$j$-th support point for
        $\partial_x^i f$ with $j\in[n_i]$};
    \draw[->] (j.west) to[bend left] (x.south);
    \node[anchor=west] (i) at (1, +1) {$i$-th derivative of $f$ with $i\in[l]$};
    \draw[->] (i.west) to[bend right] (x.north);
    \node[anchor=west] (t) at (1.5, 0) {$t$-th process execution with $t\in[T]$};
    \draw[->] (t.west) to[bend right] (x.east);
\end{tikzpicture}
\caption{Short overview of our notation.}\label{f:notation}\end{figure}

Its left to show how to generate the support points as input for
Algorithm~\ref{a:data-gen}. One way is to use support points of a real
process curve dataset, and using Algorithm~\ref{a:data-gen} to create
semi-synthetic copy of it. In a fully synthetic setting, the support points at execution
$t\in[T]$, the support points $(x^i(t),y^i(t))$ can be sampled from a
distribution on $\RR^{n_i}$ respectively, whose statistical properties
change over the temporal axis. For instance, 
$y^i(t)\sim\mathcal{N}_{\mu^i(t),\sigma}$ with $\mu^i:[T]\to\RR^{n_i}$
encoding the drift behavior over the temporal axis for the support
points. Another free parameter of
Algorithm~\ref{a:data-gen} is the function $f$ to use. In principle,
$f$ can be chosen from any parametrized function set, like B-splines,
Gaussian processes~\cite{gaussian_process}, neural
networks~\cite{universal_approx_theorem}, or Kolmogorov-Arnold
networks~\cite{kans}. In Appendix~\ref{appendix:poly-example}, we
showcase in depth an example where $f$ is a polynomial.

\begin{remark}[Multivariate data]\label{r:multivariate-data}
    Our theoretic framework extends naturally to
    multivariate time series data, where each dimension $d\in[c]$ (or signal)
    has its own function $f_d$. If they do not share their latent
    information $w_d(t)$, then Algorithm~\ref{a:data-gen} can be
    executed for each dimension individually. If they share some
    latent information, then~\qref{equ:optimization-problem} can be
    extended by summing all support point conditions for all
    $f_1,\ldots,f_c$.
\end{remark}

\begin{remark}[Profile data]\label{r:profile_data}
    Our theoretic framework is also capable to generate profile data with drifts
    holding both, drifts within a profile and drifts over executions.
    This can be obtained, for instance, by
    describing how the support points should behave in each profile
    and for subsequent profiles.
\end{remark}

\section{The temporal area under the curve}\label{s:tauc}

Different usecases require different performance metrics to evaluate
algorithms. In classification, for instance, sometimes avoiding false
positives is, sometimes avoiding false negatives.  However, when it
comes to general benchmarking classifiers somewhat independently of
their precise application in the sense to see how well their response
correlates to the actual class label, the AUC~\cite{auc} is frequently
used. However, the vanilla AUC takes samples independently of their temporal
context, that is, independent of samples from the previous and next
process execution. Thus, we construct in this section a more suitable metric to measure the predictive power
of machine learning models for process drift detection. In order to
do so, we first formalize what we understand as a process drift and
which assumptions we require. 
Let $C_1, \ldots, C_T$ be a sequence of process curves and let
$\mathcal{D}\subset[T]$ be the set of curve indices belonging to 
drifts. Our first assumption is that drifts, different from point
anomalies, appear sequentially and can be uniquely decomposed into
disjoint segments:
\begin{defn}[Drift segments]\label{d:drift-segments}
Let $\cD\subset [T]$.
Then a series of subsets $\cD_1,\ldots,\cD_k\subset\cD$ is a 
\emph{partition of drift segments} if there exists $1\le l_1< h_1<l_2<
h_2<\ldots, <l_k< h_k\le T$ such that for all $i$, we have
$\cD_i=[l_i,h_i]$ and $\cD=\cup_{i=1}^k\cD_i$.
\end{defn}
\begin{figure}[htbp]
\begin{tikzpicture}

    \node[] at (4.8,-1.1)
    {\includegraphics[width=0.35\textwidth,trim={1cm 0.5cm 1cm 1cm}, clip]{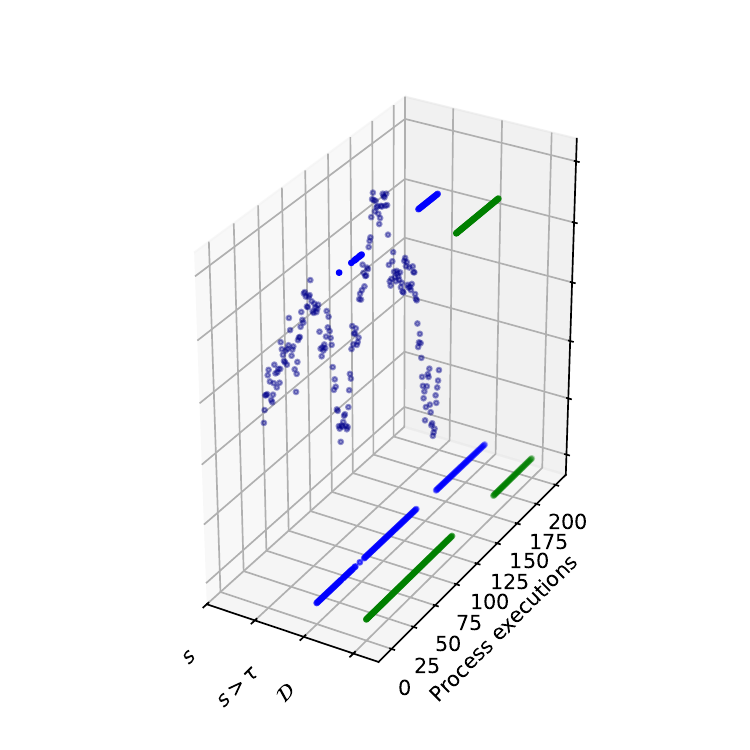}};
    \node[] at (0,-0.1)
    {\includegraphics[width=0.35\textwidth,trim={1.2cm 1cm 0.5cm 2cm}, clip]{./process_drift.pdf}};

    \draw[->, very thick] (1.5, -0.5) -- (3.1, -1);

\end{tikzpicture}
\caption{Applying a process drift detector on each process curves
yields a score $s$ which needs to be
compared to the ground
truth~$\mathcal{D}$ for each threshold $\tau$.}\label{f:detector_prediction}\end{figure}
The drift segments can be considered as a partition of the smallest
consecutive drifts which cannot decomposed any further into smaller
segments. Now, assume we also have the output $s\in\RR^T$ of a detector
where each coordinate $s_t$ quantifies how likely the curve $C_t$ of the
$i$-t-h process execution belongs to a drift, that is, the higher $s_t$ the more likely the
detector classifies $t\in\cD$ (see also Figure~\ref{f:detector_prediction}). By choosing a threshold $\tau \in \RR$,
we can construct a set
$$\hat\cD(s,\tau):=\{t\in[T]: s_t\ge\tau\}$$
which serves as a possible candidate for $\cD$. 
Clearly, if $\tau_1\ge\tau_2$, then $\hat\cD(s, \tau_1)\subseteq\hat\cD(s, \tau_2)$. 
Its also straight-forward to see that for every $\tau$, the set $\hat\cD(s, \tau)$
decomposes uniquely into
drift segments
$\hat\cD_1,\ldots,\hat\cD_l$ as defined in
Definition~\ref{d:drift-segments} and that the length and number of
these atomic segments depends on $\tau$. Now, to quantify the
predictive power of the detector yielding $s$, one needs to quantify
how \emph{close} $\hat\cD(s,\cdot)$ is to $\cD$ when $\tau$
varies. 
There are many established set-theoretic measurements that are
widely used in practice to
quantify the distance between two finite and binary sets $A$ and $B$, like the
Jaccard index $\frac{|A\cap B|}{|A\cup B|}$, the Hamming
distance
$|A\setminus B| + |B\setminus A|$, or the
Overlap coefficient $\frac{|A\cap B|}{\min(|A|, |B|)}$ just to name a
few.
Most metrics, however, have as a build-in assumption that the elements
of the set are iid and hence the temporal context is largely ignored
making them unsuitable for process drift detection. Moreover, for most
detectors we have to select a discrimination threshold $\tau$, making
evaluation cumbersome as it requires to tune the threshold on a
separate held-out dataset. Moreover, in most practical scenarios,
$\cD$ is only a small subset and thus the evaluation metric has to
consider highly imbalanced scenarios as well.

\begin{figure}[htbp]
    \centering
\scalebox{0.7}{
    \input{./overlap_score_example.tex}
}
\caption{Temporal arrangements of true and predicted drift segments as
input for Algorithm~\ref{a:overlap-score}.}
    \label{f:overlap-score}
\end{figure}
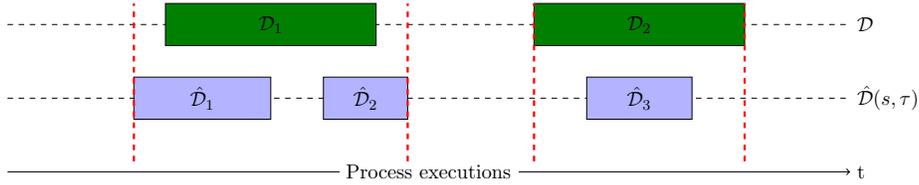

\begin{figure}[htbp]
    \includegraphics[height=0.20\textheight]{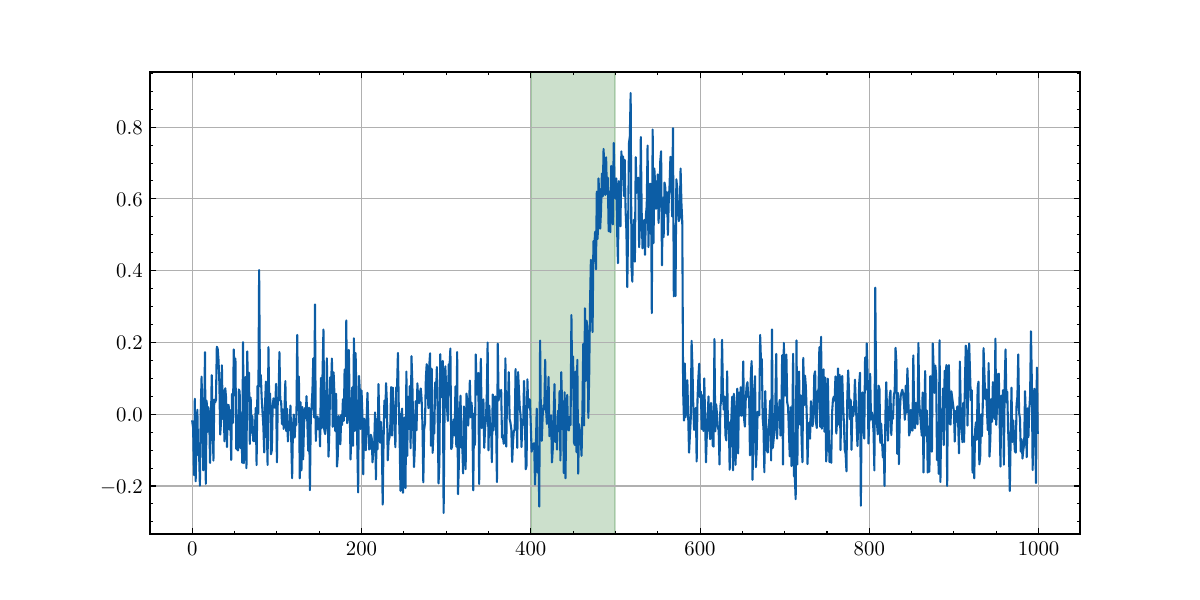}
    \includegraphics[height=0.20\textheight]{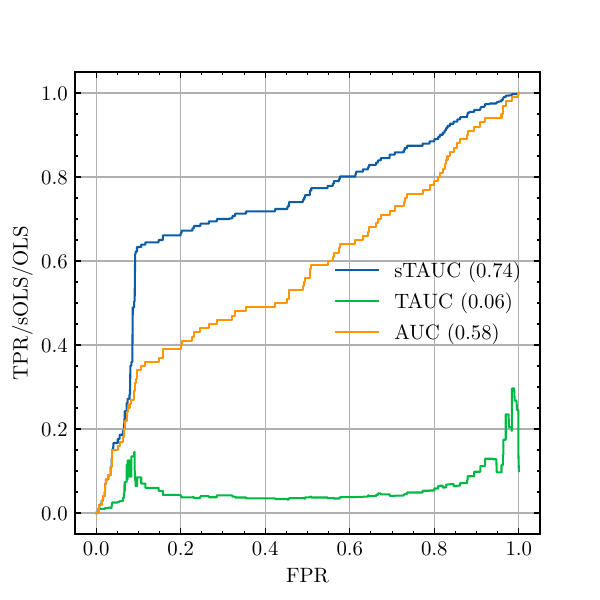}
\caption{The TPR, sOLS, and OLS when the FPR varies for the 
synthetic prediction on the left.}\label{f:tauc-stauc-auc}
\end{figure}

Clearly, detectors are required where all true drift segments $\cD_i$ are
overlapped by predicted drift segments. 
For this, let $L_i:=\{j\in[l]:
\cD_i\cap\hat\cD_j\neq\emptyset\}$. Clearly, $L_{i}\cap
L_{i+1}\neq\emptyset$ if $\cD_{i}$ and $\cD_{i+1}$ both intersect with
a predicted drift segment. Now, the set
$\cT_i:=\cup_{j\in
L_i}\hat\cD_j$ which is the union of all predictive segments
intersecting with $\cD_i$ serves as a candidate for $\cD_i$. To
measure how well $\cD_i$ is covered -  or overlapped - by $\cT_i$ we
define the \emph{soft overlap score} inspired by the Overlap
coefficient as follows:
\begin{equation}\label{equ:ols-soft}
\text{sOLS}(\cD_i, s,\tau):=\frac{|\cT_i|}{\max(\mathcal{T}_i \cup \mathcal{D}_i) -
\min(\mathcal{T}_i \cup \mathcal{D}_i) +1}
\end{equation}
Obviously, an sOLS of $1$ is the best possible and this is reached if and
only if $\cT_i=\cD_i$. It is easy to
see that for fixed $\cD_i$, the enlargement of $\cT_i$ beyond the
boundaries of $\cD_i$ improves the
overlap score, as $|\cT_i|$ increases and one of
either $\max(\cT_i\cup\cD_i)$ or $-\min(T_i\cup\cD_i)$ increases as
well. A special case is if $\cD_i$ is
completely covered by $\cT_i$, i.e. $\cD_i\subseteq\cT_i$, then
it follows that $\cT_i$ is an interval as well and thus
$\text{sOLS}(\cD_i, s,\tau)=1$.
When $\cT_i$ enlarges, then the number of false positives, i.e.
the time 
points $t$ contained in some $\hat\cD_i$ and in the complement 
$\overline{\cD}:=[T]\setminus\cD$ of the ground truth $\cD$, enlarges
as well. Thus, the predictive power of a detector is shown in the
overlap score as well as the created false positive rate
$$
\textnormal{FPR}(\cD, s, \tau):=\frac{|\hat\cD(s,\tau)\cap\overline{\cD}|}{|\overline\cD|}.
$$
into account. To also take false negatives into account, the
enumerator in~\qref{equ:ols} could be changed as follows, yielding
our final definition of the \emph{overlap score}:
\begin{equation}\label{equ:ols}
\text{OLS}(\cD_i, s,\tau):=\frac{|\cT_i\cap\cD_i|}{\max(\mathcal{T}_i \cup \mathcal{D}_i) -
\min(\mathcal{T}_i \cup \mathcal{D}_i) +1}.
\end{equation}
Algorithm \ref{a:overlap-score} illustrates in detail how the
\emph{Overlap score} $\textnormal{OLS}(\cD, s, \tau)$ can be computed
algorithmically. 
\begin{algorithm}
    \caption{Overlap score $\textnormal{OLS}(\cD, s, \tau)$}
    \label{a:overlap-score}
     \textbf{Input: }$\cD\subset[T], \; s\in\RR^T, \tau\in\RR$\hfill\textcolor{white}{.} \\
     \textbf{Output: } Overlap score\hfill\textcolor{white}{.}
    \begin{algorithmic}[1]
    \State $\cD_1,\ldots,\cD_k \gets$ find drift segments of $\cD$
    \State $\hat\cD_1,\ldots,\hat\cD_l \gets$ find drift segments of $\hat\cD(s,\tau)$
    \State $o \gets \mathbf{0} \in \mathbb{R}^k$ 
    \For{$i \in [k]$}
    \State $L_i\gets\{j\in[l]: \hat\cD_j\cap\cD_i\neq\emptyset\}$
    \Comment{All predicted drift segments overlapping with $\cD_i$}
    \State $\mathcal{T}_i\gets\cup_{j\in L_i}\hat\cD_j$ \Comment{Union
    of all segments intersecting with $\cD_i$}
        \State $o_i \gets \frac{|\mathcal{T}_i\cap\cD_i|}{\max(\mathcal{T}_i \cup \mathcal{D}_i) -
        \min(\mathcal{T}_i\cup\mathcal{D}_i)+1}$ \Comment{fraction of overlap}
    \EndFor
    \State \Return $\frac{1}{k}\sum^k_{i=1}o_i$
    \end{algorithmic}
\end{algorithm}
Our score considers both, the OLS and the FPR, which mutually
influence each other.
In the computation of the AUC, any threshold $\tau$ from
$[\min_t(s_t), \max_t(s_t)]$ yields a pair of
false positive rate $\textnormal{FPR}(\cD, s, \tau)$ and true positive
rate $\textnormal{TPR}(\cD, s, \tau)$
which can be drawn as a curve in the space where
$\textnormal{FPR}$ is on the $x$-axis and $\textnormal{TPR}$ on the
$y$-axis. 
Similarly, we define the \emph{temporal area under
the curve}, or just TAUC, as the area under the FPR-OLS curve while
the discrimination threshold $\tau$ varies. We refer to the \emph{soft} TAUC, or
just \text{sTAUC}, to the area under the FPR-sOLS curve (see
Figure~\ref{f:tauc-stauc-auc}). 

Note that the integral of the curve can be computed using two
different methods, the \emph{step rule} and the \emph{trapezoidal
rule} and
depending on which method is used, the value of the score may differ.
We showcase this behavior in detail for trivial detectors in
Appendix~\ref{appendix:trivial-tauc}. In Appendix~\ref{sec:auc-tauc},
we investigate in several synthetic cases in depth the differences and
similarities between sTAUC, TAUC, and AUC.

\begin{example}
    Consider the situation shown in Figure~\ref{f:overlap-score}.
    There, we have two true drift segments $\mathcal{D}_1$ and
    $\mathcal{D}_2$, and three segments $\hat\cD_1$, $\hat\cD_2$ and
    $\hat\cD_3$ as drift segments of some detector output $\hat\cD(s,\tau)$. 
    Clearly, $L_1=\{1,2\}$ where $L_2=\{3\}$ as only $\cD_3$
    overlaps with $\cD_2$. 
    To unveil $\cD_1$, the detector needs to separate drift segments,
    leading to false negatives and positives and thus a relatively
    small
    OLS.
      On the other hand, as $\mathcal{\hat{D}}_3 \subset
      \mathcal{D}_2$, we have 
    $\mathcal{T}_2 = \mathcal{\hat{D}}_3$. 
\end{example}

\section{Experiments}\label{s:experiments}

Next, we benchmark existing algorithms on data generated with our
framework \texttt{driftbench} and reporting the TAUC. 
All datasets and algorithms used are available in the repository
of~\texttt{driftbench}. The goal of the benchmark is to provide a
proof of concept for our score and data generation method, not to be
very comprehensive on the model side. Thus, based on
our literature research in Section~\ref{s:intro} we have
hand-selected a small set of typically used model patterns drift
detectors used in practice consists of (see
Section~\ref{s:algorithms}).

The basic evaluation loop follows a typical situation from
manufacturing, where process engineers have to identify time periods
within a larger curve datasets where the process has drifted. Thus,
all models consume as input a process curve dataset $C_1,\ldots, C_T$
and do not have access to
the ground truth $\cD$, which is the set of curves belonging to a
drift (see Section~\ref{s:data}).  Afterwards, each model predicts for
each curve $C_t$ from this dataset a score $s_t\in\RR$, and
afterwards, the TAUC, sTAUC, and AUC are computed for $s=(s_1,\ldots,s_T)$.
To account for robustness, we generate each dataset of a predefined
specification five times for a different random seed each, leading to
slightly different datasets of roughly same complexity. All models are
trained unsupervised, i.e. without any information of the true drift
segments.

\subsection{Algorithms}\label{s:algorithms}

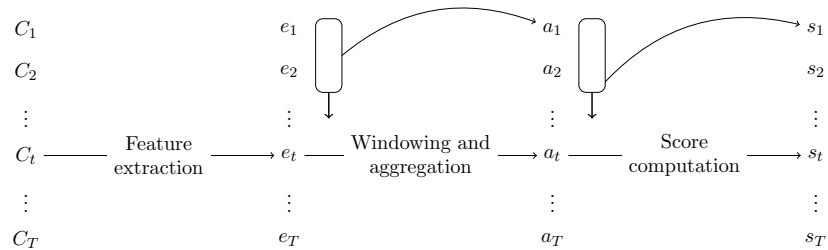
\begin{figure}[htbp]
    \scalebox{0.7}{
    \begin{tikzpicture}

        \node[](c1) at (0,0) {$C_1$};
        \node[](c2) at (0,-0.8) {$C_2$};
        \node[](c3) at (0,-1.6) {\vdots};
        \node[](c4) at (0,-2.4) {$C_t$};
        \node[](c5) at (0,-3.2) {\vdots};
        \node[](c6) at (0,-4) {$C_T$};

        \node[](l1) at (5,0) {$e_1$};
        \node[](l2) at (5,-0.8) {$e_2$};
        \node[](l3) at (5,-1.6) {\vdots};
        \node[](l4) at (5,-2.4) {$e_t$};
        \node[](l5) at (5,-3.2) {\vdots};
        \node[](l6) at (5,-4) {$e_T$};

        \node[](ll1) at (10,0) {$a_1$};
        \node[](ll2) at (10,-0.8) {$a_2$};
        \node[](ll3) at (10,-1.6) {\vdots};
        \node[](ll4) at (10,-2.4) {$a_t$};
        \node[](ll5) at (10,-3.2) {\vdots};
        \node[](ll6) at (10,-4) {$a_T$};

        \node[](s1) at (15,0) {$s_1$};
        \node[](s2) at (15,-0.8) {$s_2$};
        \node[](s3) at (15,-1.6) {\vdots};
        \node[](s4) at (15,-2.4) {$s_t$};
        \node[](s5) at (15,-3.2) {\vdots};
        \node[](s6) at (15,-4) {$s_T$};

        \draw[rounded corners] (5.5, 0.2) rectangle (6, -1.2) {};
        \draw[->, thick] (5.75, -1.2) -- (5.75, -1.7);
        \draw[->] (6, -0.5) to[bend left] (ll1);

        \draw[rounded corners] (10.5, 0.2) rectangle (11, -1.2) {};
        \draw[->, thick] (10.75, -1.2) -- (10.75, -1.7);
        \draw[->] (11, -1) to[bend left] (s1);

        \draw[->] (c4) to node[midway, fill=white, align=center]{Feature\\extraction} (l4);
        \draw[->] (l4) to node[midway, fill=white, align=center]{Windowing and \\aggregation} (ll4);
        \draw[->] (ll4) to node[midway, fill=white, align=center]{Score\\computation} (s4);

    \end{tikzpicture}
}
    
\caption{A high-level overview of the elementary tasks of the
detectors used.}\label{f:high-level}\end{figure}

The algorithms used can be decomposed into multiple steps (see also
Figure~\ref{f:high-level}), but not all algorithms use all steps.
First, there are may some features extracted from each curve.
Afterwards, a sliding window collects and may aggregate these such
that a score is computed. 

\subsubsection{Feature extraction}

In this step, we use autoencoders~\cite{autoencoder} to
compute a $k$-dimensional representation $e_t\in\RR^k$ for each
high-dimensional process curve $C_t$ with $k$ small. The indention behind is to
estimate an inverse of the unknown function $f$ and to
recover information about the support points used.
Moreover, we also apply deterministic aggregations over the
$x$-information of each curve $C_t$.

\subsubsection{Windowing and aggregation}

In this step, the algorithms may aggregate the data from the previous
step using a
fixed window of size $m$ that is applied in a rolling fashion along the
process iterations.
One aggregation we use is to first compute for each
coordinate $j\in[k]$ of $e_t\in\RR^k$ with $t\ge m$ the rolling mean
$a_{t, j} = \frac{1}{m} \sum^t_{i=t-m+1} e_{i, j}$.
These values can then further be statistically aggregated, like by
taking the maximum
$a_t:=\max\{a_{t,j}: j\in[k]\}$.

\subsubsection{Score computing}

Goal of this step is to compute a threshold which correlates with the
ground truth, that is, the larger the higher the possibility of a
drift. Here, we may also aggregate previous features in a rolling
fashion. The simplest aggregation we use is to compute the euclidean
distance of subsequent elements
$s_t = \|a_t - a_{t-1}\|_2$ which is just the absolute difference if
$a_t$ and $a_{t-1}$ are scalars. If $a_t$ is a scalar, we also can
compute the rolling standard deviation, again over a window of size
$m$, like this:
$$s_t = \sqrt{\frac{1}{m-1}\sum^t_{j=t-m+1}\left(a_j -
\left(\frac{1}{m} \sum^t_{i=t-m+1} a_i\right)\right)^2}.$$

Another approach follows a probabilistic path by testing if a set of
subsequent datapoints $\{a_{t-m+1},\ldots, a_t\}$ come from the same distribution as a given reference
set. In our study, we use a windowed version~\cite{kswin} of the popular Kolmogorov-Smirnov
test~\cite{ks_test}, often called KSWIN, which makes no assumption of the
underlying data distribution. However, this can
only be applied when $a_t$ is a scalar. More particularly, we define two window sizes, $m_r$  for the
reference data and $m_o$ for the observation. The windows are offset by constant
$\delta > 0$. We then invoke the KS-test and receive a $p$-value
$p_t$, which is small if the datasets come from different
distributions. Thus, one way to derive a final score is to compute 
$s_t=\log(1+\frac{1}{p_t})$.
Another probabilistic method we use in our study based on a multivariate 
statistical test is the \emph{Maximum Mean Discrepancy} (MMD)~\cite{mmd}.
This method uses feature mappings based on kernels, and calculates the distance between the means 
in these mappings. MMD also makes no assumption about the underlying distribution, and works on
multidimensional data. We use this method in the same way using two windows as described in the KS-test.
We also evaluate algorithms that derive their score based on a
similarity search within $\{a_1,\ldots, a_t\}$. 
Here, we use clustering
algorithms, like the popular $k$-means algorithm, and use the euclidean distance to the computed
cluster center of $a_t$ as $s_t$. Another way is
to fit a probability density function on $s_t$, like a mixture of
Gaussian distributions, and to set $s_t$ as the log
likelihood of $a_t$ within this model.

\subsection{Algorithm Overview}

Here is a short summary of the algorithms used in our benchmark study:

\begin{itemize}
    \item{\texttt{RollingMeanDifference($m_r$)}
	   First, the rolling mean over a window of size $m_r$ is computed
	   over all values for the respective curves in the window. Afterwards, the maximum
	   value for each curve is taken and the absolute difference between two consecutive
  maximum values is computed.}
    \item{\texttt{RollingMeanStandardDeviation($m_r$)}
	   First, the rolling mean over a window of size $m_r$ is computed over all values
	   for the respective curves in the window. We also choose the maximum value of these
	   computed values per curve. Then, we compute the standard deviation using the same window
	   for this one-dimensional input.}
    \item{\texttt{SlidingKSWIN($m_r, m_o, \delta$)}:
	   We compute the mean value for each curve and apply a sliding KS-test on this
	   aggregated data. We use two windows of size $m_r$ and $m_o$ where the windows are offset by 
           $\delta$.}
    \item{\texttt{Cluster($n_c$)}: A cluster algorithm performed on the raw
        curves using $n_c$ clusters where score is distance to the closest
    cluster center.}
    \item{\texttt{AE($k$)-mean-KS($m_r, m_o, \delta$)}: First, an autoencoder is applied
            extracting computing $k$ many latent dimensions.
            Afterwards, the mean across all $k$ latent dimensions is computed. 
            Finally, a sliding KS-test is applied with two windows 
	    of sizes  $m_r$ and $m_o$, where the windows are offset  by $\delta$.}
    \item{\texttt{AE($k$)-MMD($m_r, m_o, \delta$)}: First, an autoencoder is applied
            extracting computing $k$ many latent dimensions.
            Afterwards, a $k$-dimensional sliding MMD-test is applied with two windows 
	    of sizes  $m_r$ and $m_o$, where the windows are offset by $\delta$.}
\end{itemize}

\subsection{Datasets}\label{s:data}

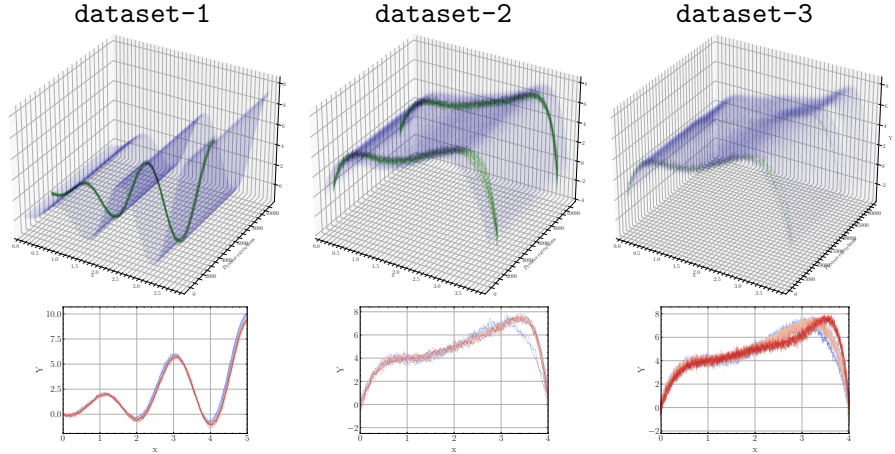
\begin{figure}[htbp]
    \centering
    \begin{tikzpicture}
   \node[] (d1) at (0, 0) {\includegraphics[width=0.27\textwidth]{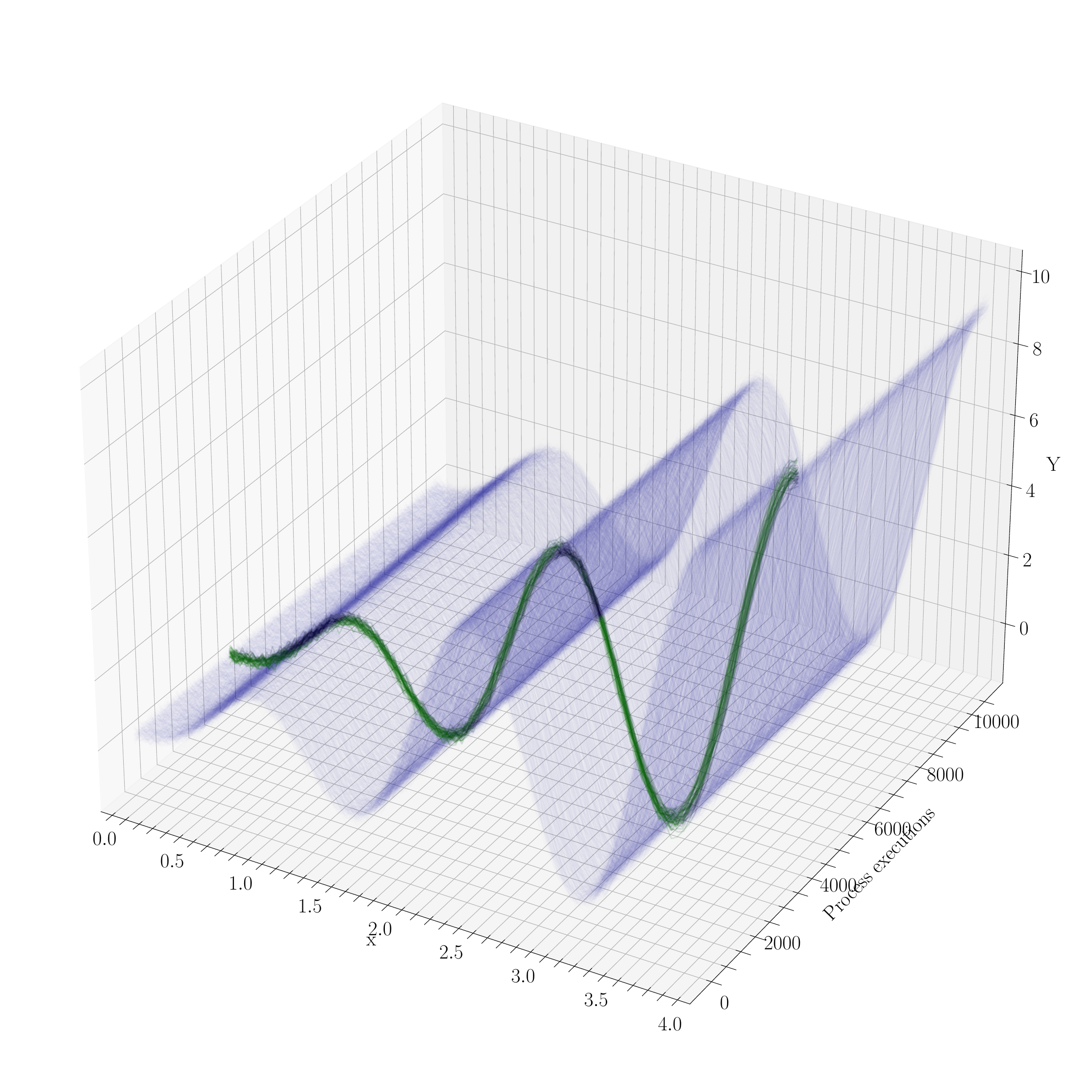}};
   \node[] (d2) at (4, 0) {\includegraphics[width=0.27\textwidth]{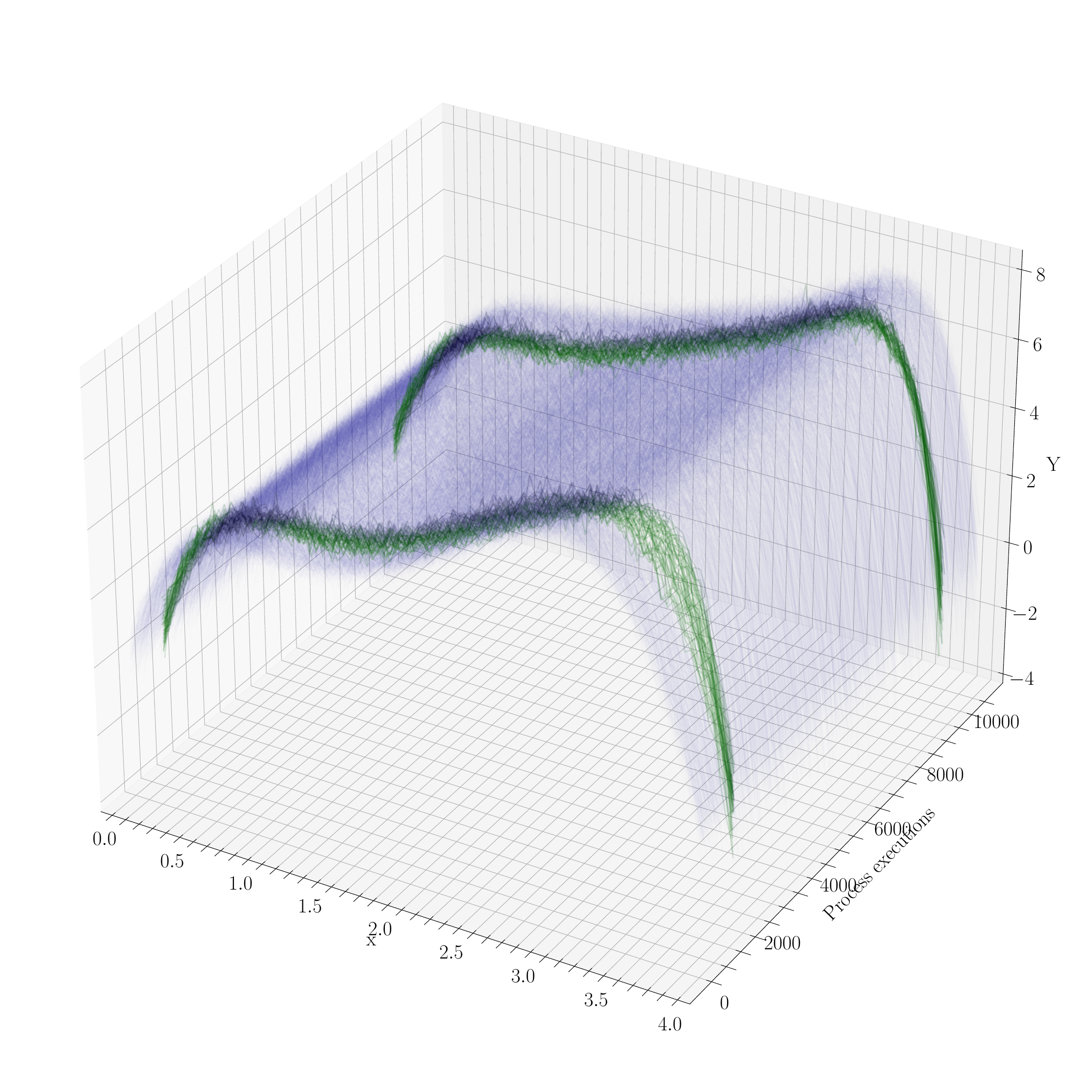}};
   \node[] (d3) at (8, 0) {\includegraphics[width=0.27\textwidth]{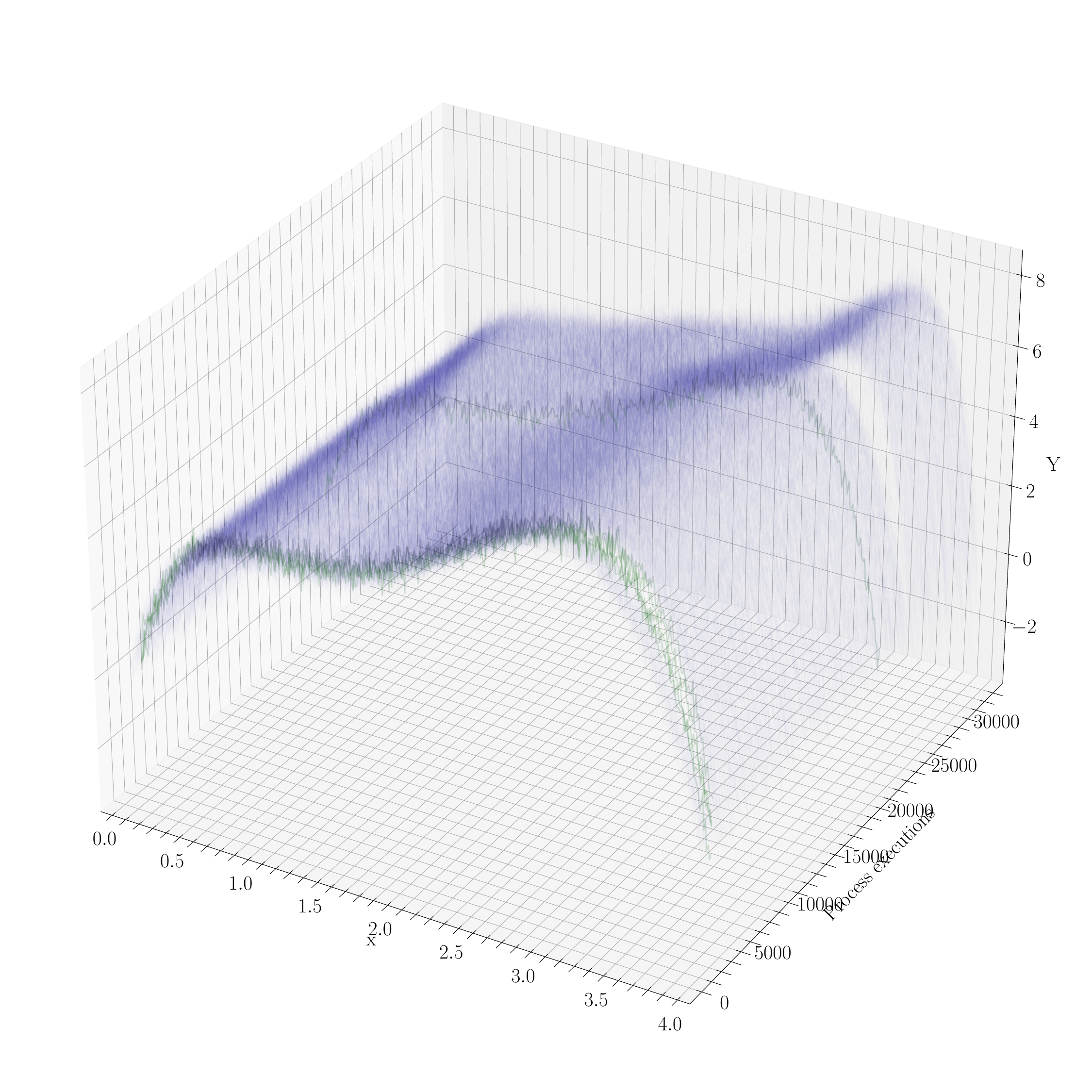}};

   \node[] at (0, -2.9) {\includegraphics[width=0.2\textwidth]{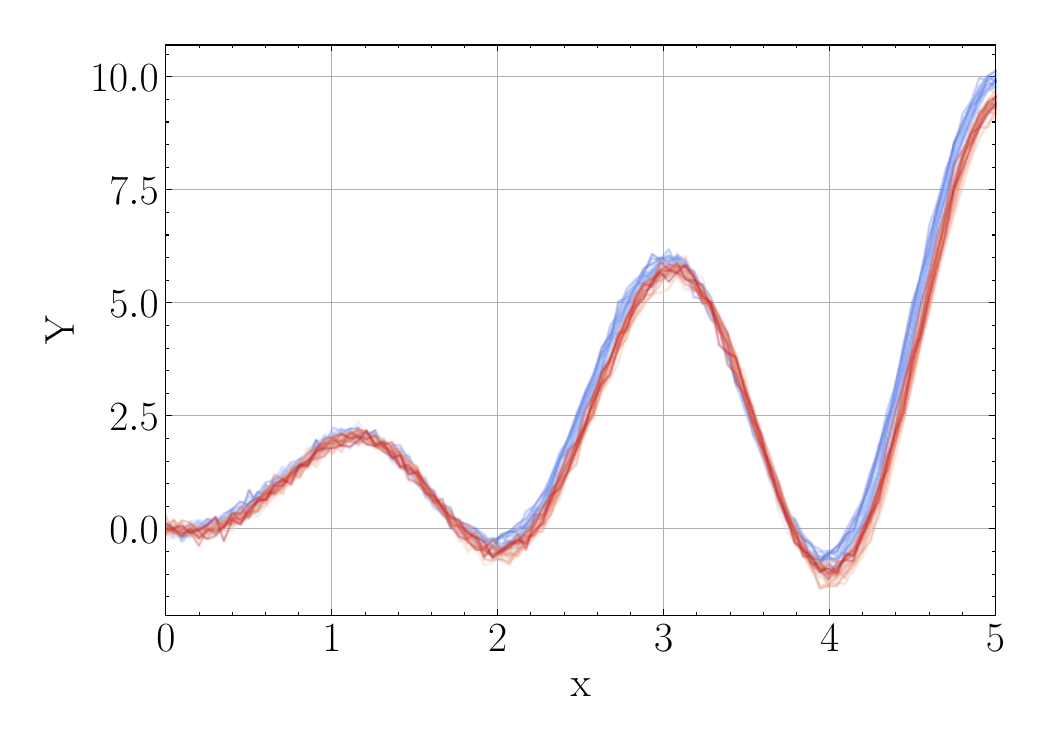}};
   \node[] at (4, -2.9) {\includegraphics[width=0.2\textwidth]{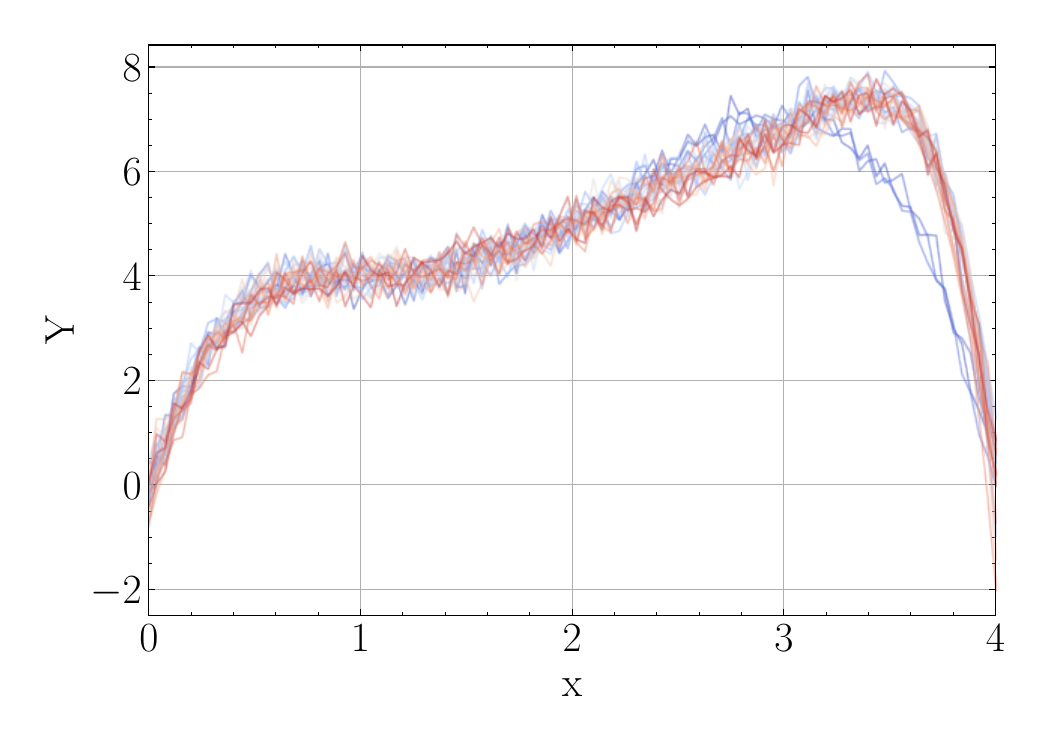}};
   \node[] at (8, -2.9) {\includegraphics[width=0.2\textwidth]{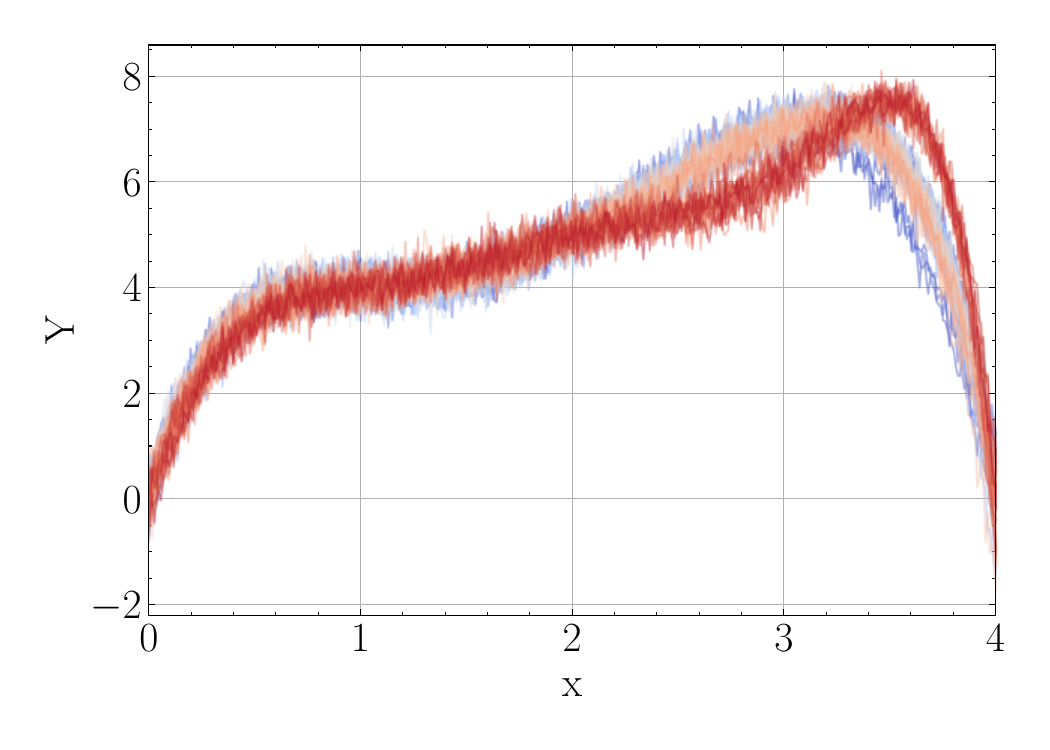}};

   \node[above=-0.5cm of d1] {\Dlargemsingdrift};
   \node[above=-0.5cm of d2] {\Dlargemuldrift};
   \node[above=-0.5cm of d3] {\Dextrememuldrift};
    \end{tikzpicture}
\caption{The datasets used in our benchmark study. The true drift
    segments are marked in green. Lower figures show selected curves,
    whose color encodes the process iteration $t\in[T]$ - blue marks
    smaller $t$ values, red larger ones. Recall that $\texttt{dataset-$k$}$ has $k$ many
drift segments. 
}\label{f:data-large-multiple-drift}
\end{figure}

We
benchmark the algorithms listed in~\ref{s:algorithms} on three
different datasets (see Figure~\ref{f:data-large-multiple-drift})
created with our framework \texttt{driftbench}, all designed to
comprise different inherent challenges. 
The datasets \Dlargemuldrift{} and \Dextrememuldrift{} have been
inspired by the force signals of staking processes (see
also~\cite[A.1]{causalassembly}) where we used
$f(w, x)=\sum_{i=0}^7w_i\cdot x^i$
as function to generate them.  The \Dlargemuldrift{} consists of $T=10.000$
curves, each called on $|I]=100$ equidistant values between $[0,4]$,
i.e., $\overline{x}=0$ and $\Delta x=0.04$. On the other hand,
\Dextrememuldrift{} consists of $T=30.000$ curves each having $|I|=400$
values between $[0,4]$. Both datasets have drifts that concern a movement of the
global maximum together with drifts where only information of first order
changes over time. In the generation process of
\Dlargemsingdrift{}, we used 
$f(w,x)=w_0\cdot x\cdot \sin(\pi\cdot x-w_1)+w_2\cdot x$
and generated $T=10.000$ many curves, each having $|I|=100$
datapoints. It only holds a single drift, where the global minimum at
drifts consistently over a small period of time along the $x$-axis.
In all datasets, the relative number of curves belonging to a drift is very
small: roughly $1$ percent in \Dlargemsingdrift{}, $2$ percent in \Dlargemuldrift{}, and $0.1$ percent
in \Dextrememuldrift{}. Particularly, $\texttt{dataset-$k$}$ has $k$ many
drift segments.
To generate a drift segment $[t_0, t_1]$ for a given support point
where the value should change linearly from $a$ to $b$ (see also
Section~\ref{appendix:poly-example}), we
sampled from normal distributions $\mathcal{N}_{\mu(t),\sigma}$ with fixed
$\sigma$ and mean
$$
\mu(t)=
\begin{cases}
    a, &\textnormal{ if } t < t_0\\
    b\cdot \frac{t-t_0}{t_1-t_0}+a, &\textnormal{ if } t_0\le t\le t_1\\
    b, &\textnormal{ if } t_1 < t
\end{cases}.$$

\subsection{Results}
\begin{figure}[htbp]
\centering
    \includegraphics[width=\textwidth]{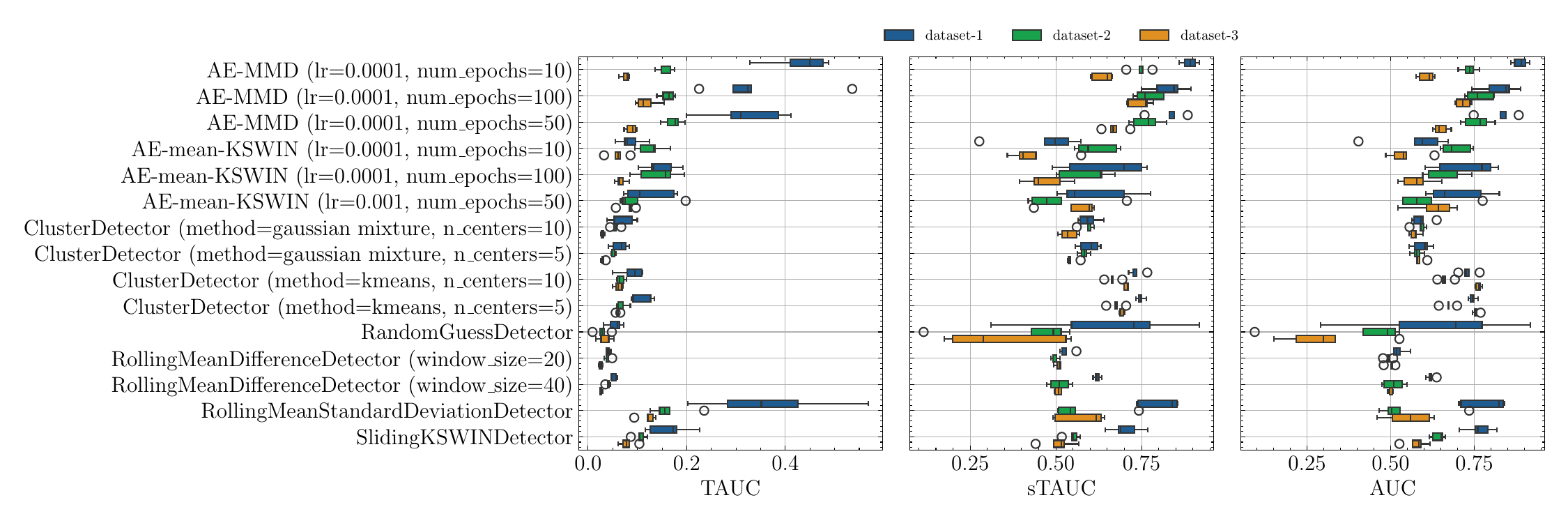}
    \caption{Benchmark results on \Dlargemsingdrift{},
        \Dlargemuldrift{}, and
        \Dextrememuldrift{}.}\label{f:benchmark}
\end{figure}

\begin{figure}[htbp]
\centering
    \includegraphics[width=\textwidth]{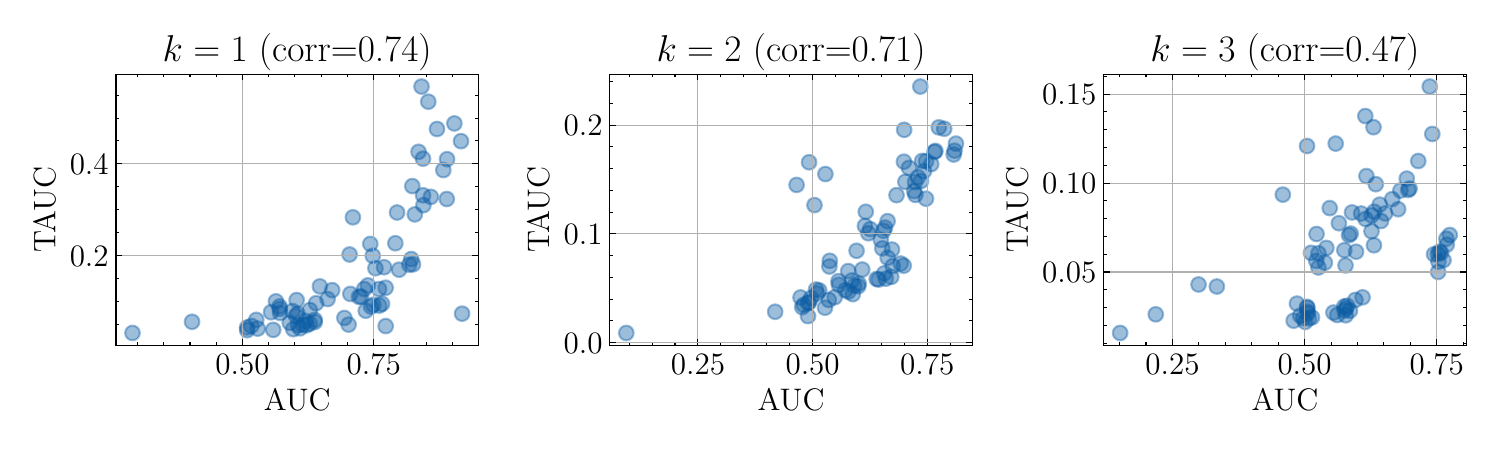}
    \caption{Correlation between TAUC and AUC for different number of
    true drift segments $k$.}\label{f:benchmark_tauc_vs_auc}
\end{figure}

The result of our benchmark study is shown in
Figure~\ref{f:benchmark}. Generally, there is a discrepancy in
detectors of the highest AUC and the highest TAUC. More concrete, the larger the
number of true drift segments in a dataset is, the larger the
discrepancy (see also
Figure~\ref{f:benchmark_tauc_vs_auc}). 
For instance,
the \texttt{RandomGuessDetector} reached the highest AUC
score on \Dlargemsingdrift{}, where it ranges on all three datasets
among the last ranks in the TAUC score. 
On all datasets, autoencoder-based systems reach among the best
detectors for both, TAUC and AUC. Those using a multivariate test
in their latent space reach better scores than these using an
aggregation of multiple uni-variate tests. 
Although some
cluster-based systems archive good AUC scores on 
dataset~\Dextrememuldrift{}, none of the benchmarked algorithms is
capable to compute a score that can be used to recover the true
drift segments, resulting in small TAUC scores for all algorithms (see
also Figure~\ref{f:best-detectors3}). The respective predictions
over the temporal dimension of the best
detectors are shown in Appendix~\ref{appendix:best_detectors} in more
detail, where it also becomes visible that detectors with higher TAUC
better recover the true drift segments.


\section{Conclusion}

This work shows how algorithms designed to detect process drifts can
be benchmarked in robust and reliable way. We have introduced a scalable and controllable data
generation method that creates process curves datasets with drifts and
a verified ground truth. In our approach, process curve datasets can
be solely generated by modelling the behavior of support points over
the temporal axis and using non-linear optimization.
We then introduce and study the novel
\emph{TAUC} score which is
particularly designed to evaluate the performance of drift detectors
on their temporal consistency over sequential process executions.
We proved the effectiveness of our approach in a small
benchmark study. Our results reveal that
existing algorithms often struggle with datasets containing multiple
drift segments, underscoring the need for further research. 

\subsection*{Acknowledgements}
This work is supported by the Hightech Agenda Bavaria.  The authors
are grateful to Matthias Burkhardt, Fabian Hueber, Kai Müller, and
Ulrich Göhner
for helpful discussions. We also thank the anonymous referees for
helpful comments and suggestions.

\subsection*{Data availability}
The generated data and implemented algorithms are implemented in a
python package \texttt{driftbench} which is freely available under
\url{https://github.com/edgarWolf/driftbench}. 

\subsection*{Conflict of interests}
The authors provide no conflict of interest associated with the
content of this article.

\bibliographystyle{ieeetr}
\bibliography{drift}

\newpage
\appendix

\section{Predictions of detectors of benchmark study}\label{appendix:best_detectors}

\begin{figure}[H]
    \centering
    \includegraphics[width=0.9\textwidth]{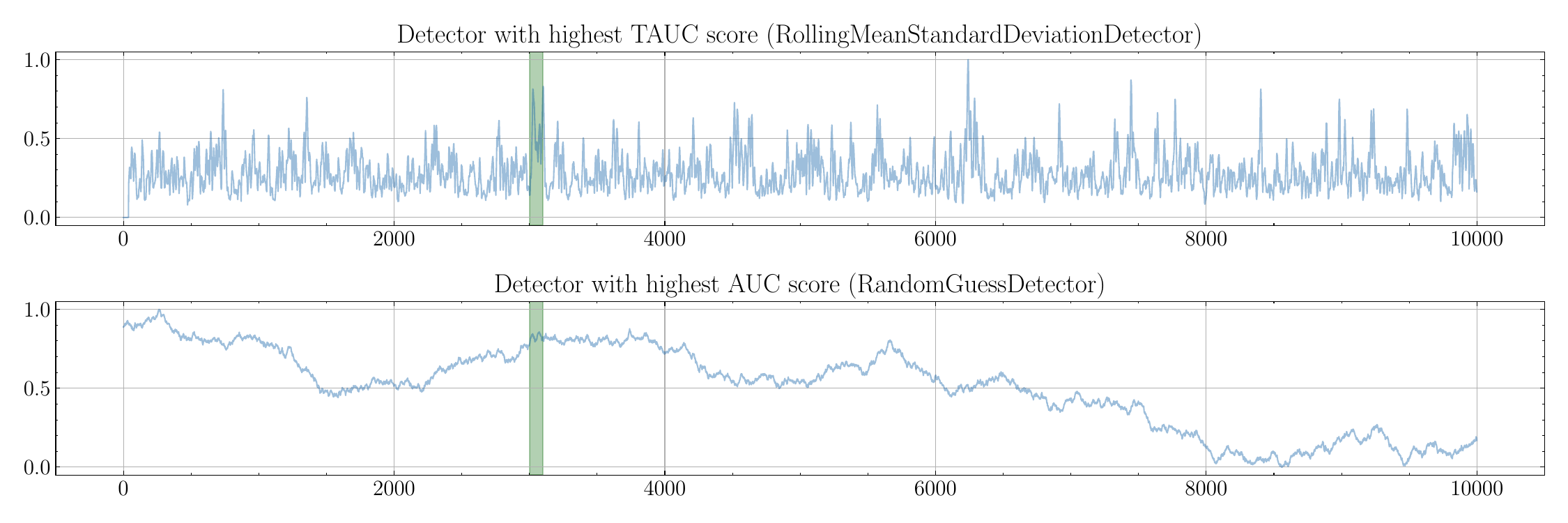}

\caption{Best detectors on \Dlargemsingdrift{}.}\label{f:best-detectors1}
\end{figure}

\begin{figure}[H]
    \centering
    \includegraphics[width=0.9\textwidth]{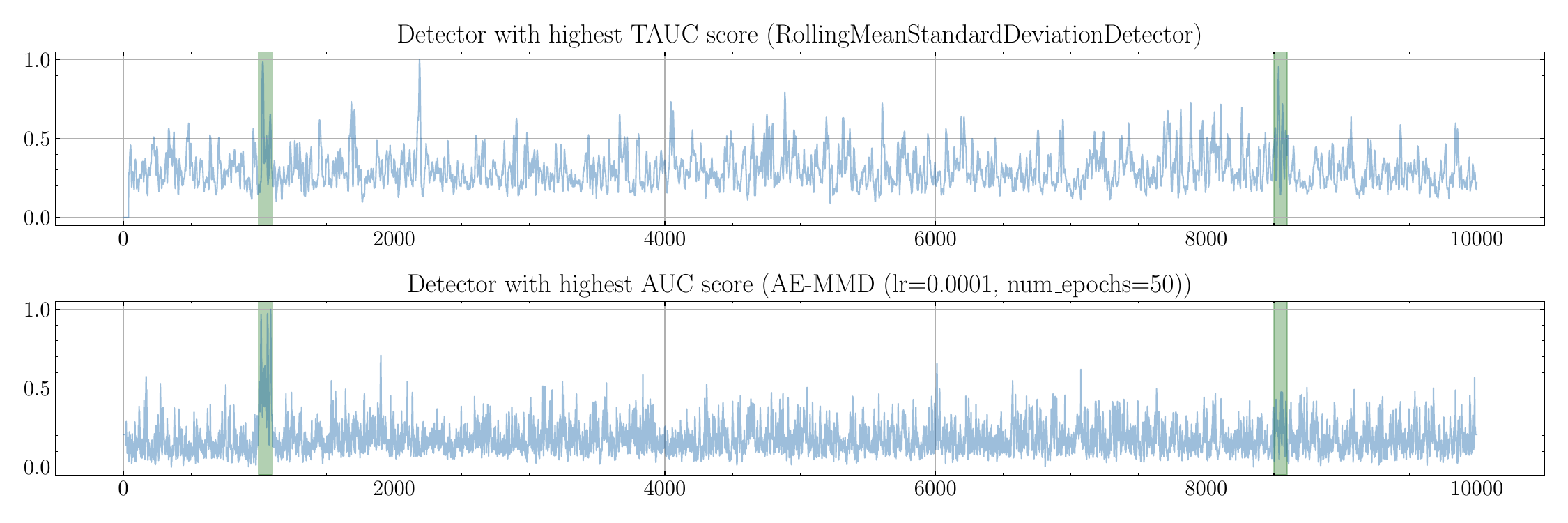}

\caption{Best detectors on \Dlargemuldrift{}.}\label{f:best-detectors2}
\end{figure}

\begin{figure}[H]
    \centering
    \includegraphics[width=0.9\textwidth]{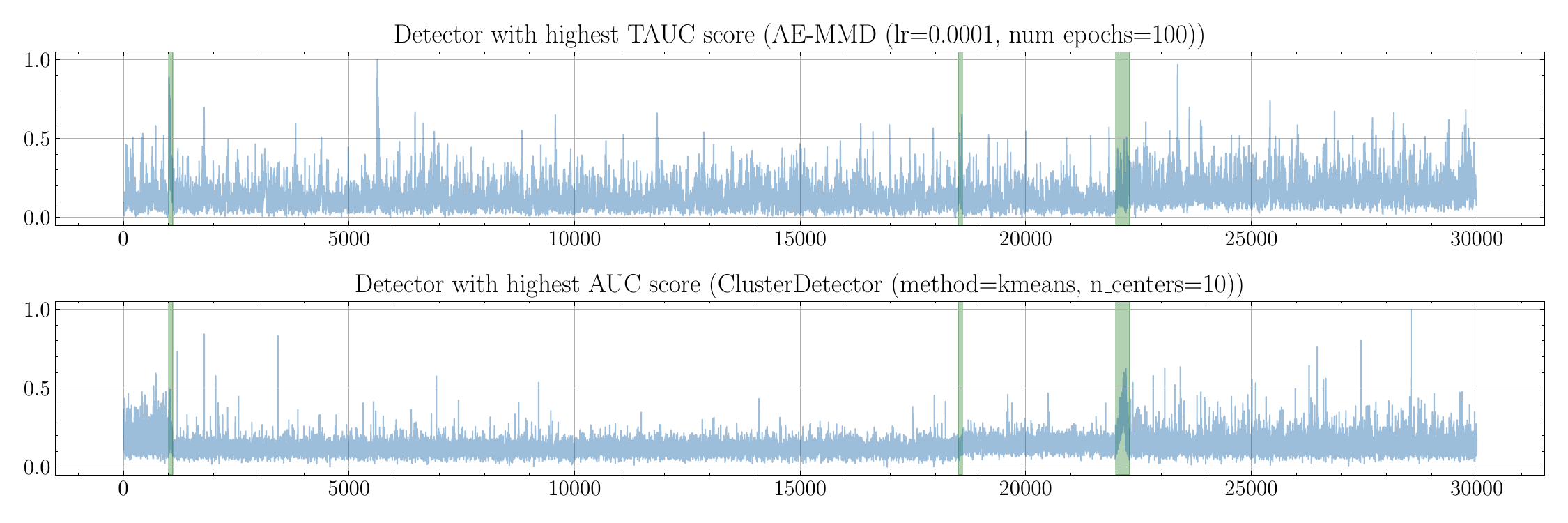}

\caption{Best detectors on \Dextrememuldrift{}.}\label{f:best-detectors3}
\end{figure}

Figure~\ref{f:best-detectors1},
Figure~\ref{f:best-detectors2}, and Figure~\ref{f:best-detectors3}
show
the 
prediction $s\in\RR^T$ of the detectors reaching highest TAUC and AUC scores on
the individual datasets are shown respectively.

\section{Data generation with polynomials}\label{appendix:poly-example}

\begin{figure}[H]

    \centering
       \begin{tikzpicture}
\node[](t1) at (0,0) {\includegraphics[width=0.3\textwidth]{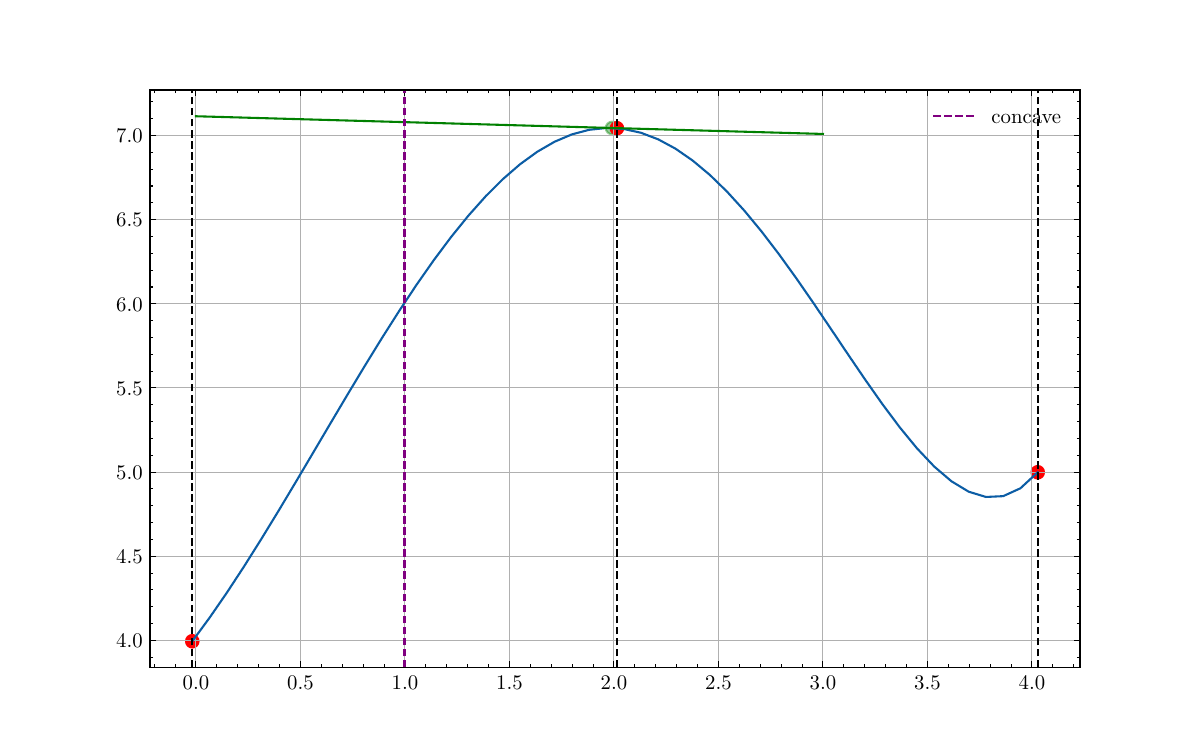}};
\node[](t2) at (5,0) {\includegraphics[width=0.3\textwidth]{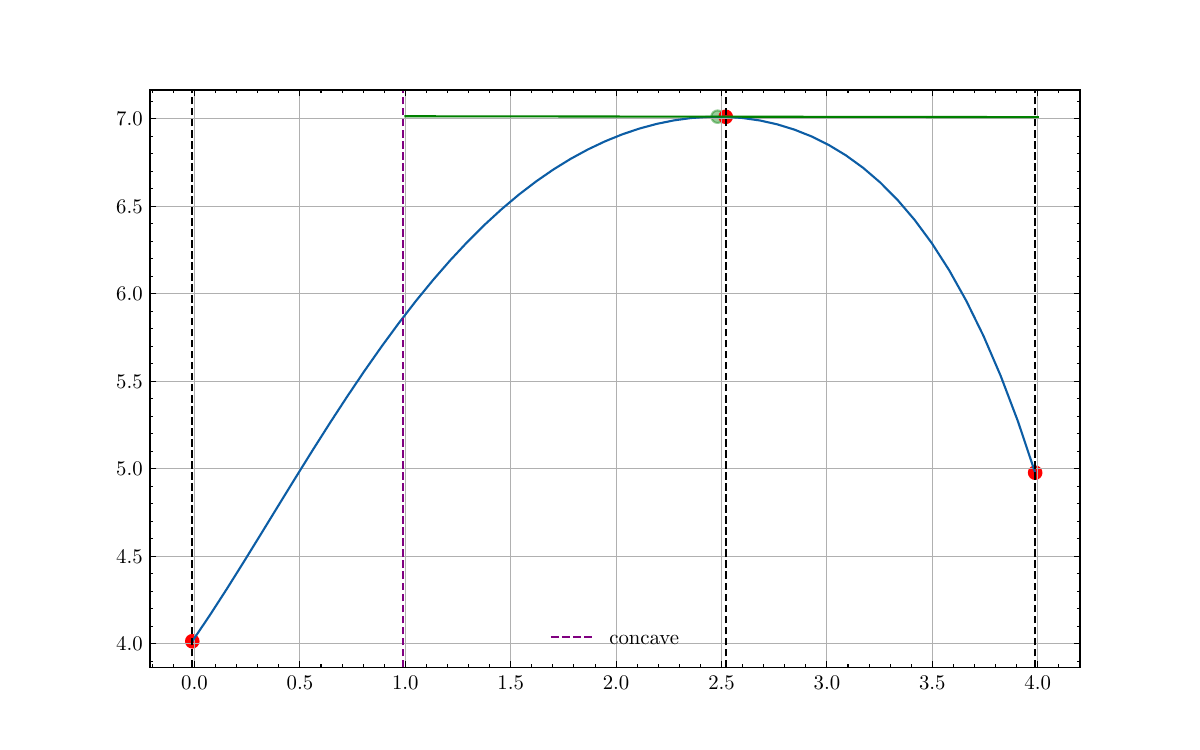}};
\node[](t3) at (10,0) {\includegraphics[width=0.3\textwidth]{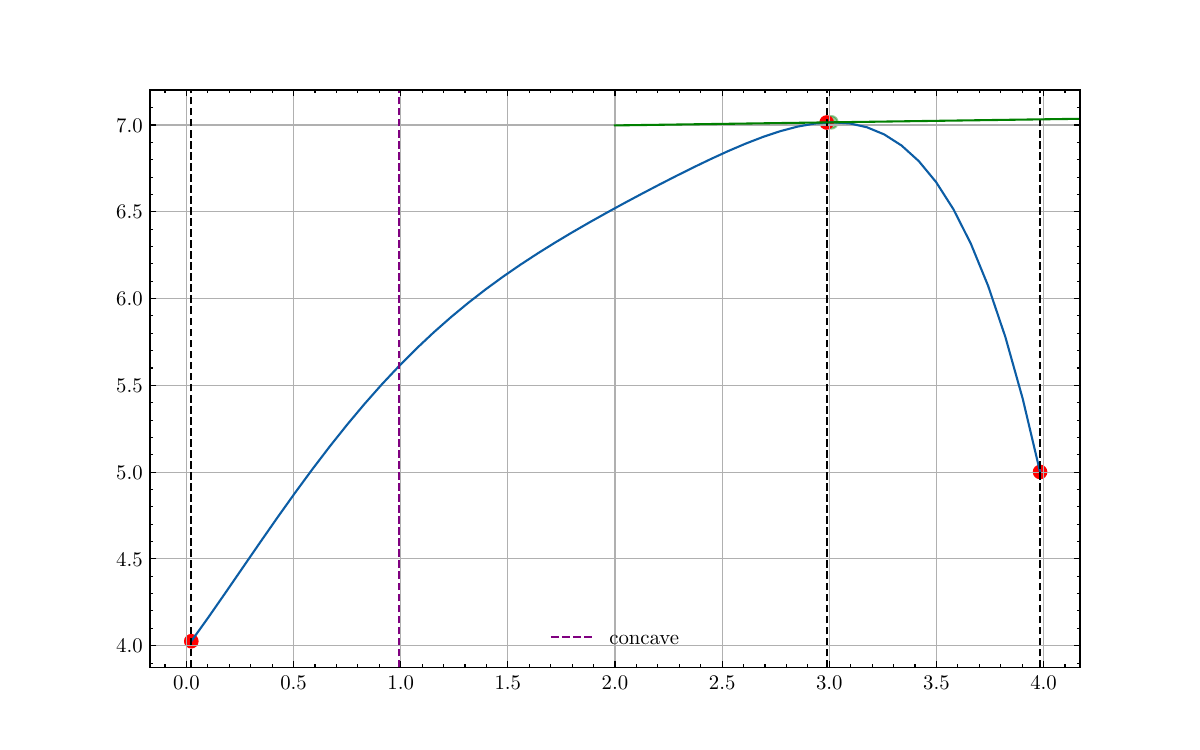}};

   \node[below=-0.2cm of t1] {$t=1000$};
   \node[below=-0.2cm of t2] {$t=1250$};
   \node[below=-0.2cm of t3] {$t=1300$};
       \end{tikzpicture}

    \caption{Visualization of some process curves in the example
        dataset. The red dots indicate support points with first order
        information given.
    The green line visualizes the slope at the green dot, encoded by
    the condition for the first derivative. The purple dashed line indicates the curvature at the
    corresponding $x$-value, encoded by the condition for the second derivative.
    From $t=1000$ to $t=1300$, the $x$-value of the maximum moves from
    $2$ to $3$.}
\label{f:curves_drift}\end{figure}
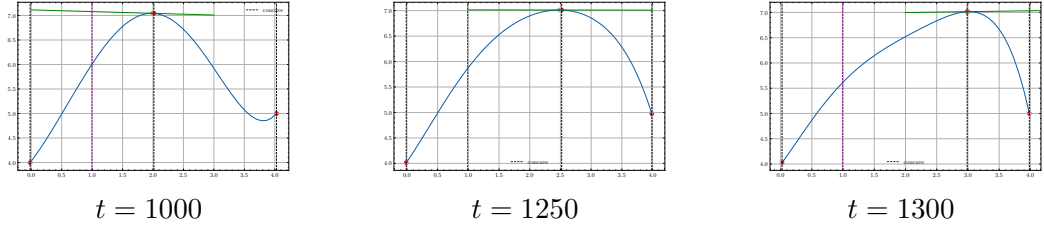

In this section, we demonstrate the data generation method
introduced in Section~\ref{s:data-generation} along an example
involving a polynomial $f:\RR^6\times\RR\to\RR$ of degree five, i.e. 
$$f(w, x)=\sum_{i=0}^5w_i\cdot x^i$$
with $w\in\RR^6.$
We simulate $2000$ process executions and thus sample $2000$ process curves. 
The shape of each curve is defined by its support points. We are only
interested in its curvature in $I=[0, 4]$. First, we want to add
a condition onto the start and end of the interval, namely that $f(w,
0)=4$ and $f(w,4)=5$. Moreover, we
would like to have a global maximum at $x=2$, which means the first
order
derivative 
$$\partial^1_x f(w, 2)=\sum_{i=1}^4 i\cdot w_i\cdot 2^{i-1}$$ 
should be zero and its second order derivate 
$$\partial^2_x f(w, 2)=\sum_{i=1}^3 i\cdot(i-1)\cdot w_i\cdot 2^{i-2}$$ 
should be smaller than
zero. Here, we want it to be $-1$. Finally, we want to the curve to be
concave at around $x=-1$.
All in all, these conditions result into the following equations, some
of them are visualized in Figure~\ref{f:curves_drift}:
\begin{align*}
    \partial^0_x f(w, 2) &= 7 & \partial^1_x f(w, 2) &= 0 & \partial^2_x f(w, 2) &= -1\\ 
     \partial^0_x f(w, 0) &= 4    & \partial^0_x f(w, 4) &= 5 & \partial^2_x f(w, 1) &= -1 
\end{align*}
Then, we let the data drift at some particular features. We simulate a scenario, 
where the peak at $x^0_1$ and $x^1_0$ moves from the $x$-position $2$ to $3$ during the process 
executions $t = 1000$ until $t = 1300$. Thus, we let $x^0_1$ and $x^1_0$ drift from $2$ to $3$,
resulting in a change of position of the peak.
We let the corresponding $y$-values $y^0_1 = 7$ and $y^1_0 = 0$ unchanged. 
Now, we can solve each of the $2000$ optimization problems, which results in $2000$ sets of
coefficients for each process curve, such that the conditions are satisfied. By evaluating $f$
with the retrieved coefficients in our region of interest $[0, 4]$, we get $2000$
synthesized process curves with a drift present at our defined drift segment from 
$t = 1000$ until $t = 1300$. 

\begin{figure}[H]
\centering
\begin{tikzpicture}
        \node[] at (0,0) {\includegraphics[width=0.35\textwidth]{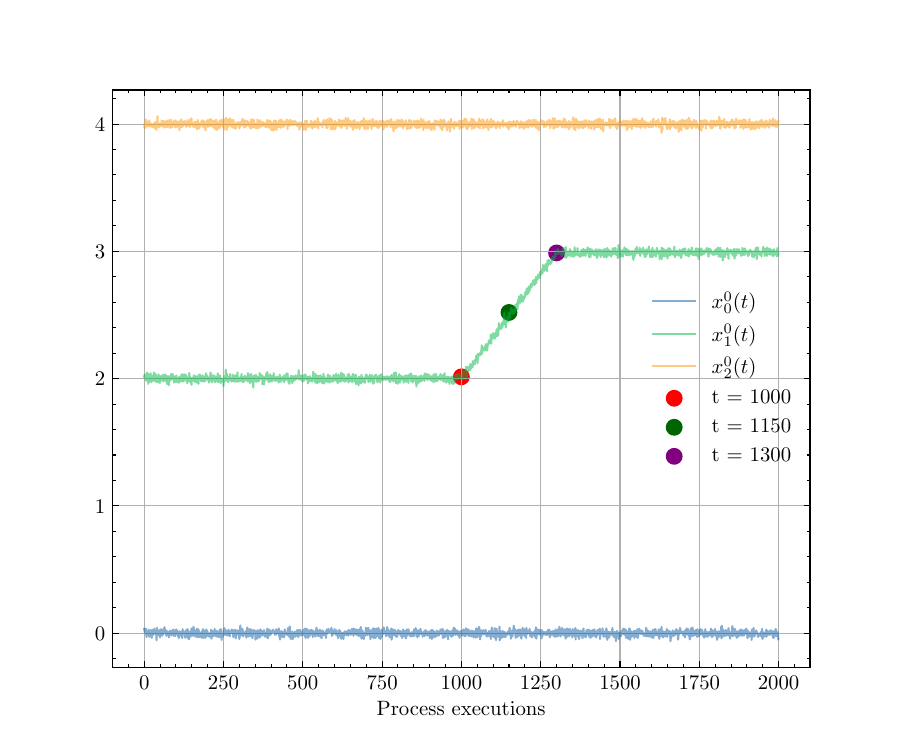}};
        \node[] at (5,0) {\includegraphics[width=0.35\textwidth]{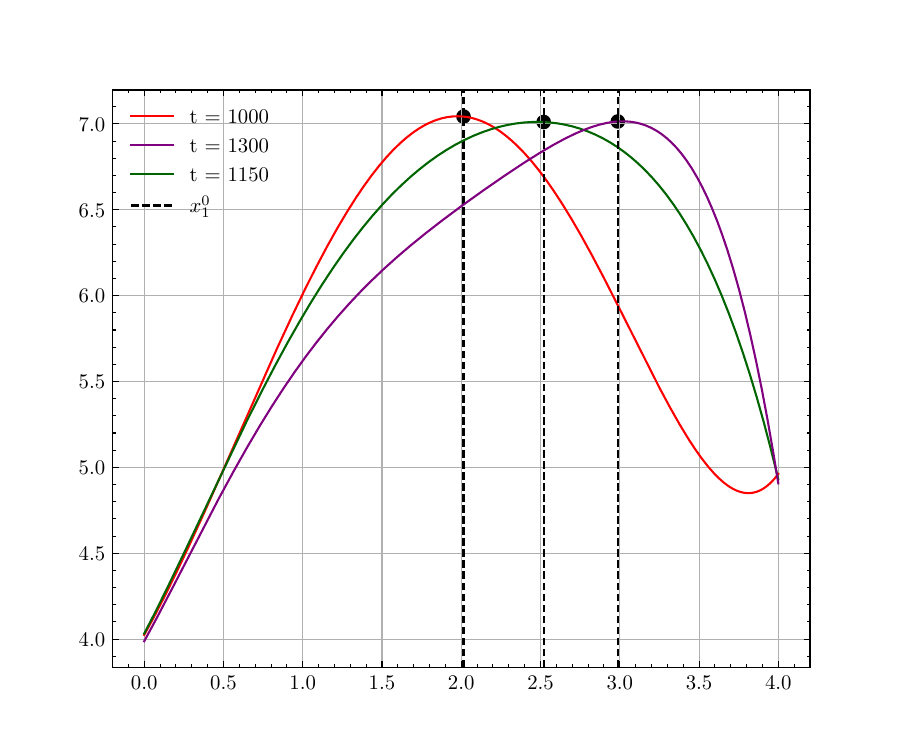}};
        \node[] at (10,0) {\includegraphics[width=0.35\textwidth]{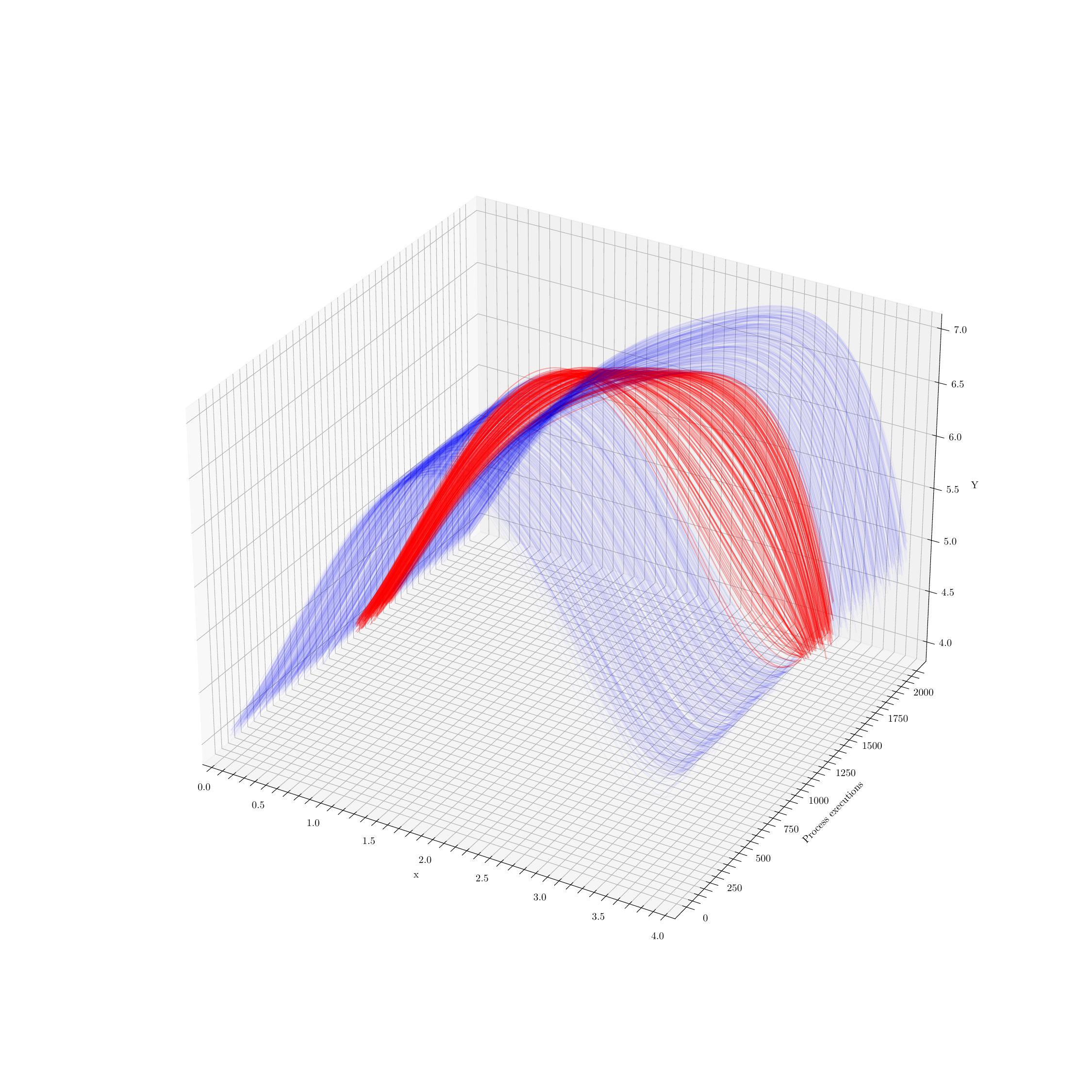}};
    \end{tikzpicture}
    \caption{Visualization of the drift applied on $x^0_1$ in this example, with respective curves. The left figure
    shows how the $x^0_i$ values change over time. Only $x^0_1$ changes, as by our drift definition from the process
    executions $t = 1000$ until $t = 1300$ linearly from $2$ to $3$, the others remain unchanged.
    The middle figure shows the respective curves, color-coded to the dots in the left figure. 
    } 
\end{figure}

\section{TAUC vs AUC}\label{sec:auc-tauc}

In this section we explore in depth the similarities and differences
of the TAUC introduced in Section~\ref{s:tauc} and the
established AUC. This is done along synthetic predictions. 

\subsection{Lagged prediction}\label{sec:auc-tauc-lagged}

\begin{figure}[H]
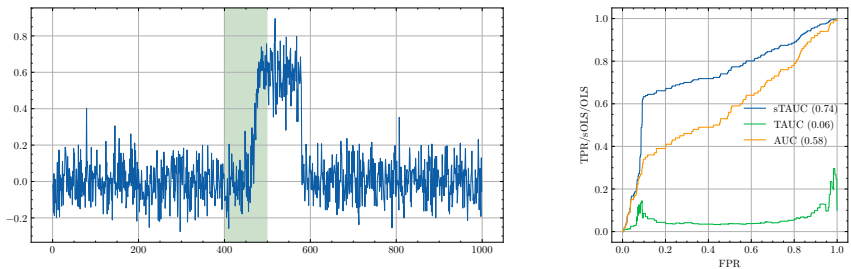

    \includegraphics[height=0.20\textheight]{./tauc_vs_auc_delayed.pdf}
    \includegraphics[height=0.20\textheight]{./tauc_vs_auc_delayed_curves.pdf}
\caption{Prediction of a detector that lags behind the ground truth
(left) and its curves underneath the TAUC and AUC (right).}\label{f:tauc-auc-delayed}\end{figure}

The first example we look at is a typical scenario that appears if
window-based approaches are used, namely that the prediction lags a
bit behind of the true window, but still the detector overlaps a
significant proportion of the drift segment (see
Figure~\ref{f:tauc-auc-delayed}. Other than the TPR, the sOLS rewards these
predictors and thus the sTAUC shows a larger value than the AUC.

\subsection{Change point detection}

\begin{figure}[htbp]
    \includegraphics[height=0.20\textheight]{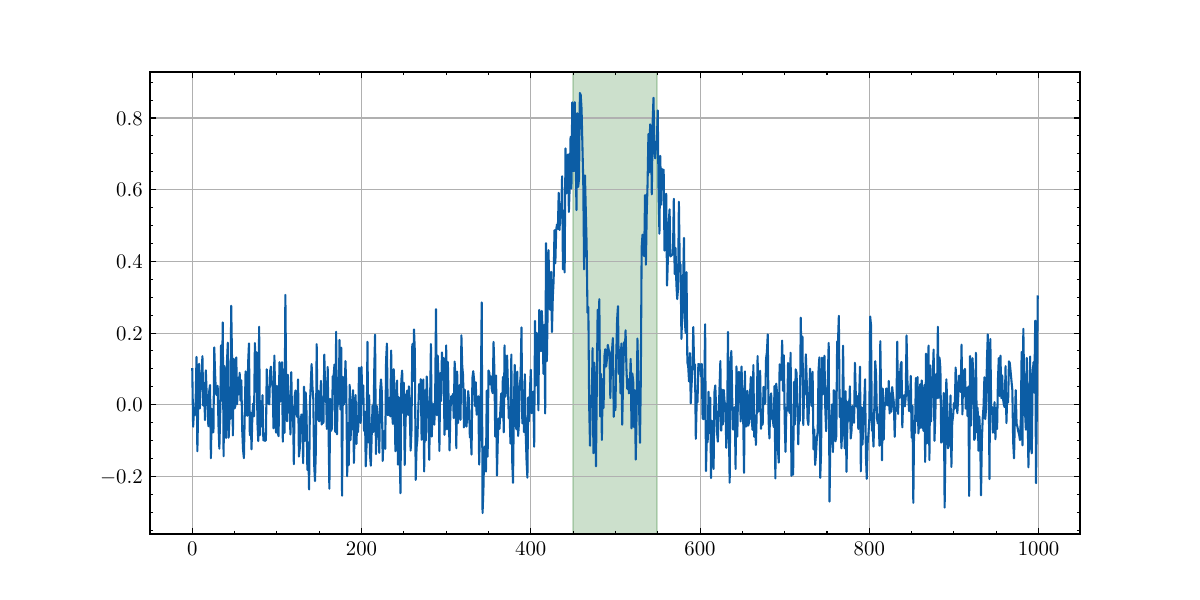}
    \includegraphics[height=0.20\textheight]{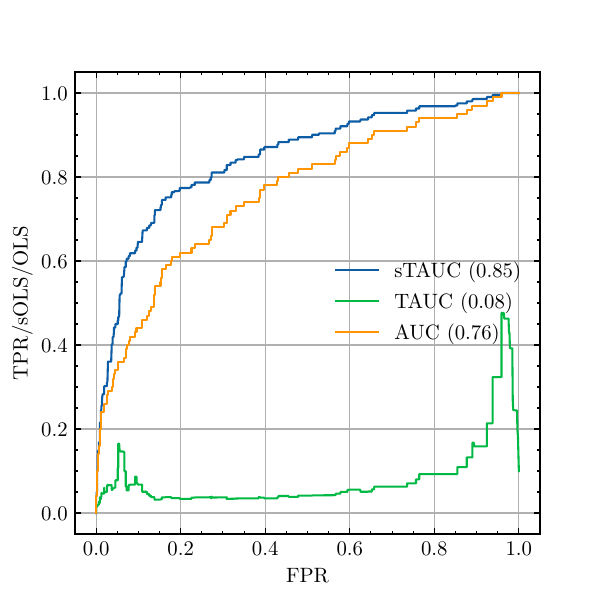}
\caption{Prediction of a detector that shows high scores at the
    boundary of the true drift segment only
(left) and its curves underneath the TAUC and AUC (right).}\label{f:tauc-auc-cp}\end{figure}

Another typical scenario is that a detector shows significantly large
values at the start and end of the true drift segment, but sag
in between (see Figure~\ref{f:tauc-auc-cp}). This could appear when
using methods based on detecting change points. In principal, the
detector correctly identifies the temporal context of the drift
segment, although showing lower scores while the curves drift. Such
predictions also score higher values in the sTAUC than the AUC.

\subsection{Varying length and position of predicted segments}

\begin{figure}[H]
    \includegraphics[height=0.20\textheight]{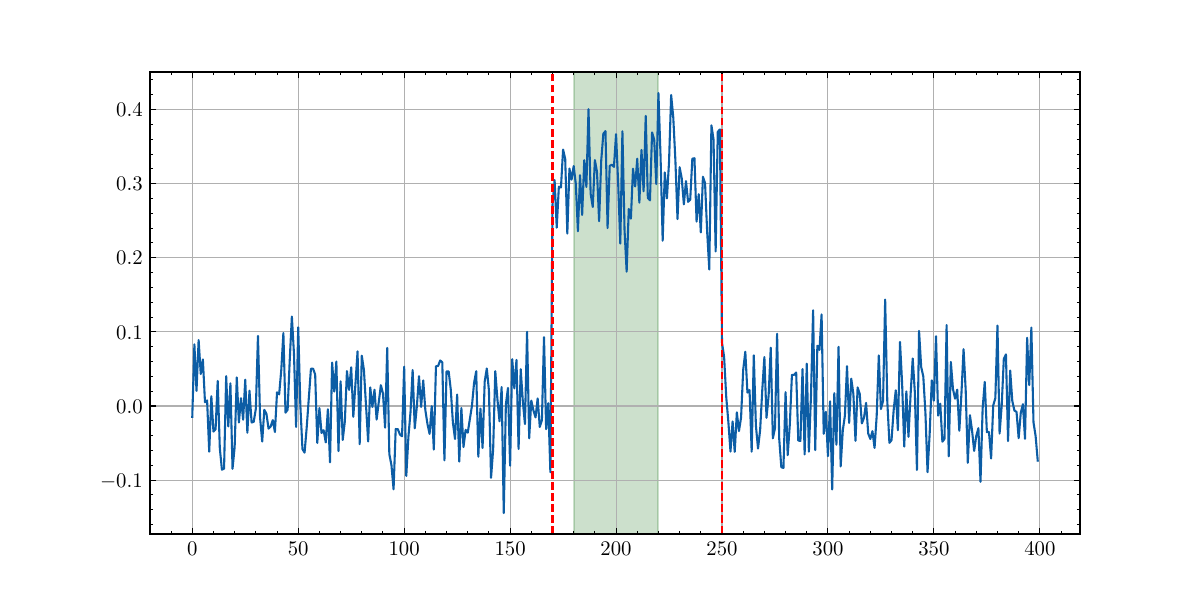}
    \caption{Situation with single predicted segment (red dashed) and single true
segment (green area).}\label{f:tauc-auc-moving}\end{figure}

\begin{figure}[H]
    \includegraphics[width=0.45\textwidth]{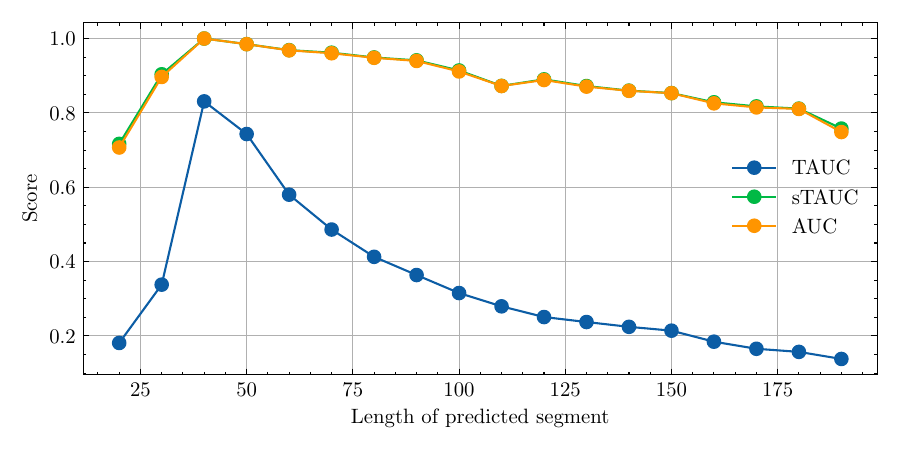}
    \includegraphics[width=0.45\textwidth]{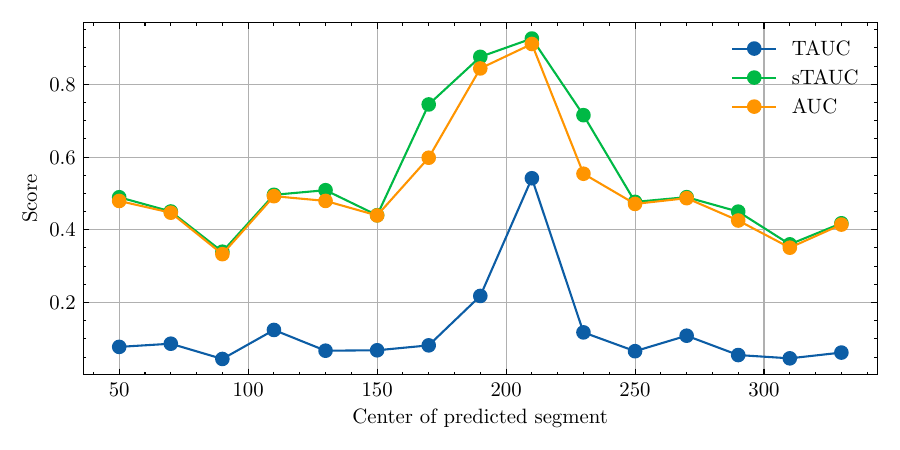}
\caption{Behavior of sTAUC, TAUC, and AUC when length and
position of predicted segment varies.}\label{f:tauc-auc-moving-score}\end{figure}

A situation where the sTAUC coincides with the AUC mostly is
in when only one true and predicted drift segment exist (see
Figure~\ref{f:tauc-auc-moving}).  In cases where the center of the
predicted segment coincides with the center of the true segment, the
AUC and sTAUC match almost exactly when the length of the predicted
segment is varied (see left graphic in
Figure~\ref{f:tauc-auc-moving-score}). If the predicted segment has
fixed length that equals the length of the true segment and the
position of its center is varied from $50$ to $350$, AUC and sTAUC
coincide mostly, but the sTAUC shows a faster rise when the predicted
segment overlaps with the true segment due to the effects explained in
Section~\ref{sec:auc-tauc-lagged}.

\section{TAUC for trivial detector}\label{appendix:trivial-tauc}

To get a better understanding of the TAUC, we showcase the behavior
on trivial detectors based on the structure of the ground truth.
Suppose two pair of points $(\text{FPR}_i, \text{OLS}_i)$ and $(\text{FPR}_{i+1}, \text{OLS}_{i+1})$ of the constructed curve. 
Then the two methods for computing the TAUC are the following:
\begin{itemize}
	\item \textbf{Trapezoidal rule:} \\
	Construct the curve by linearly interpolating $\text{OLS}_i$ and $\text{OLS}_{i+1}$ 
	in between $\text{FPR}_i$ and $\text{FPR}_{i+1}$ and then calculate the area 
	under the curve by using the trapezoidal integration rule.
        \item \textbf{Step rule:} \\
	Construct the curve by filling the values in between $\text{FPR}_i$ and 
	$\text{FPR}_{i+1}$ with a constant value of $\text{OLS}_i$ and then 
	calculate the area under the curve by using the step rule.
\end{itemize}
For example, take the trivial detector that always predicts a drift,
called \texttt{AlwaysGuesser}. 
Then we receive the two points $(0, 0)$ and $(1, \frac{P}{k})$ as the only two points of the curve,
where $P$ denotes the portion of drifts in $y$ and $k$ denotes the number of drift segments in $y$.
In case of the step function, the computed score will always be $0$,
since the constructed curve only contains one step from $[0, 1)$ with a OLS-value of $0$, 
and only reaches a OLS-value of $\frac{P}{k}$ when reaching a FPR of $1$ on the $x$-axis. Hence, 
the area under this constructed curve is always $0$.
When using the trapezoidal rule, we linearly interpolate the two obtained trivial points of the 
curve, thus constructing a line from $(0, 0)$ to $(1, \frac{P}{K})$. The TAUC is then given by 
the area under this line, which is equal to $\frac{P}{2k}$.
Now suppose a detector which never indicates a drift, called \texttt{NeverGuesser}.
Then we receive $(0, 0)$ as our only point, which does not construct a curve and thus does not have an area under it. Hence, the TAUC for this trivial detection is $0$ in both cases. 

\begin{figure}[htbp]
    \centering
    \begin{tikzpicture}
        \node[] at (0,0) {\includegraphics[width=0.45\textwidth]{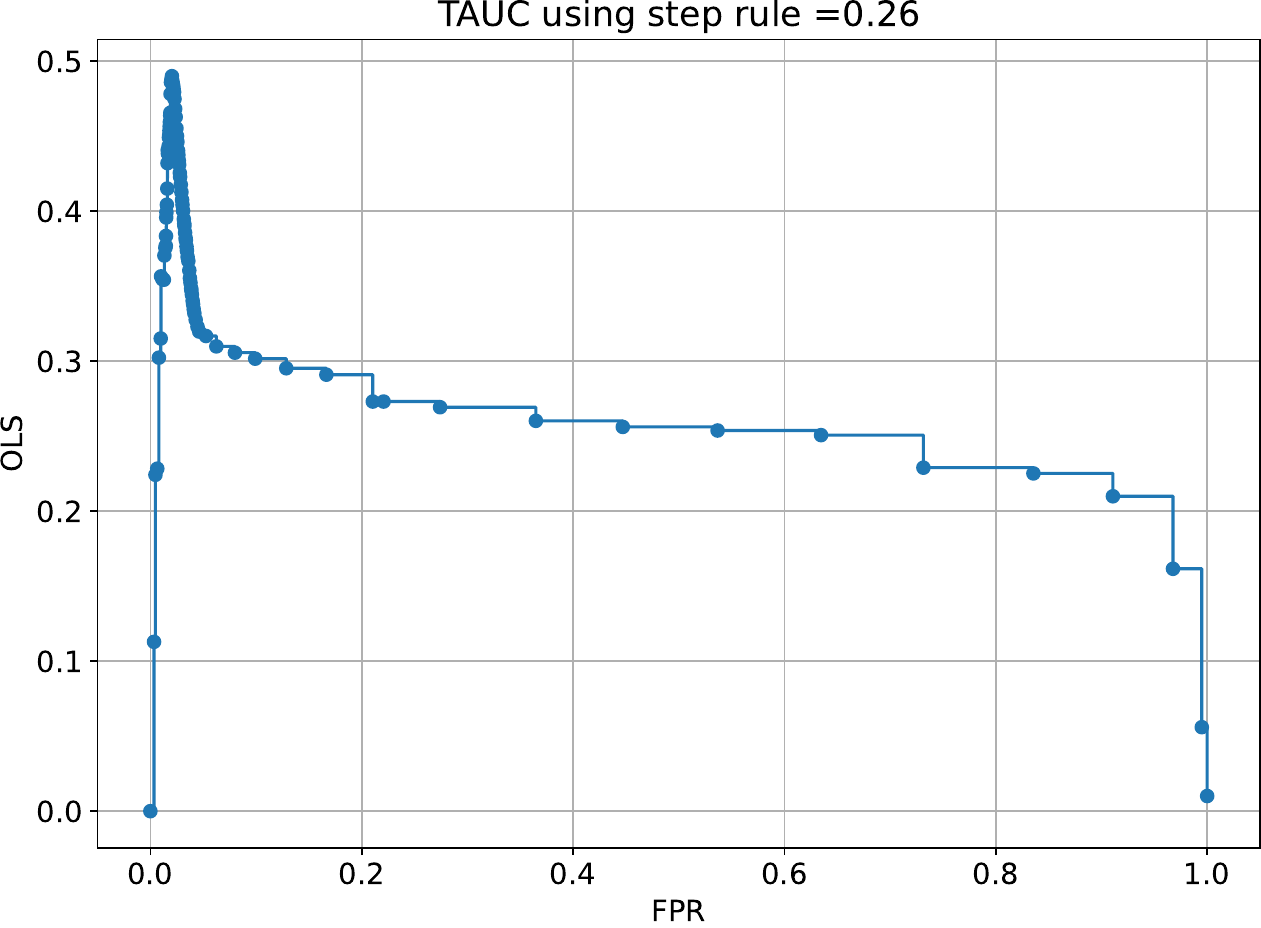}};
        \node[] at (7,0) {\includegraphics[width=0.45\textwidth]{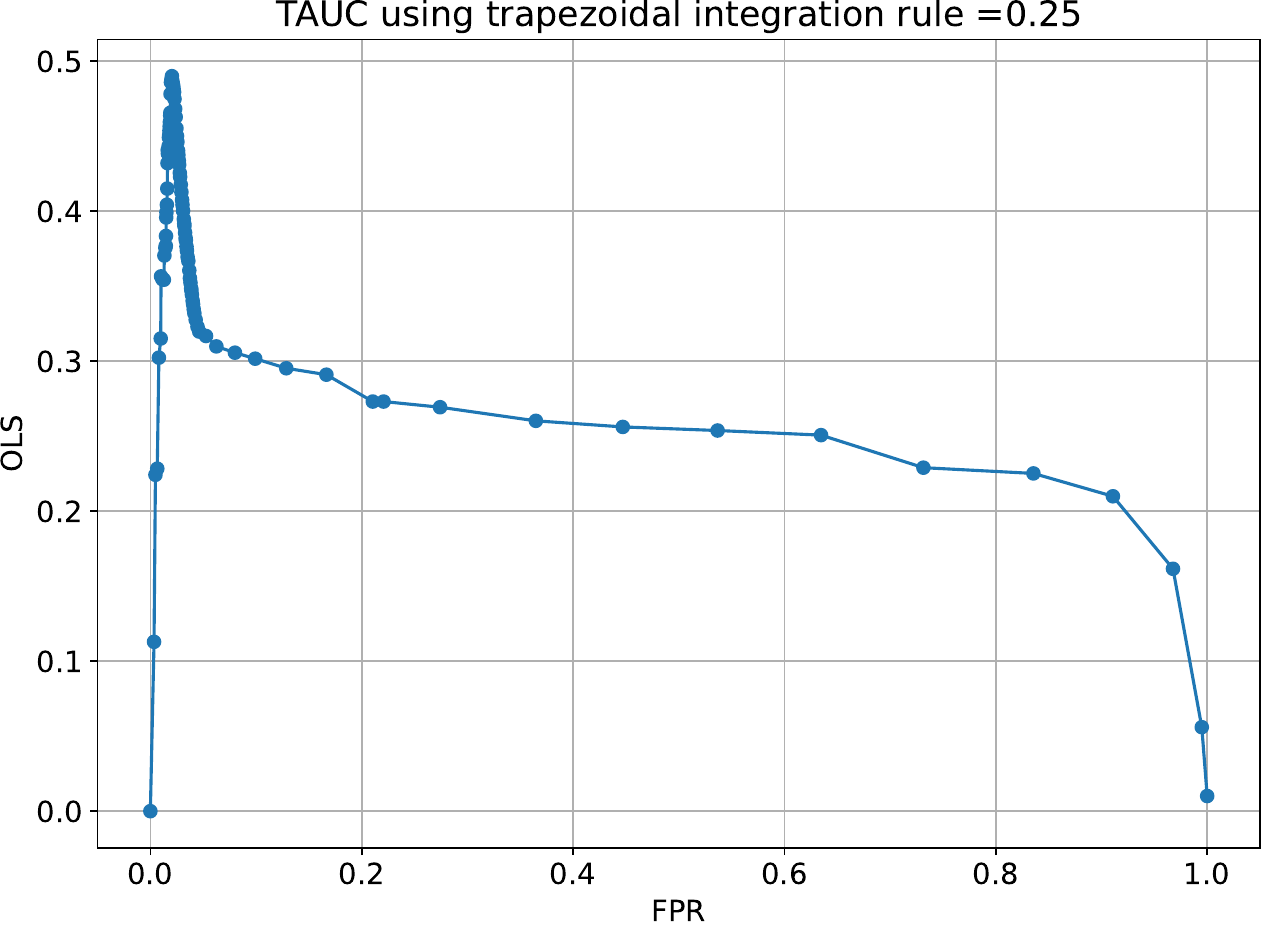}};
        
    \end{tikzpicture}
    \caption{Visualization of a concrete curve used to calculate the TAUC with its TAUC-score. 
    Left figure shows the constructed curve when using the step rule, 
    while the right figure shows the curve when calculating the TAUC using the 
    trapezoidal integration rule.}
\end{figure}

In order to investigate how the TAUC behaves with an increasing number of segments 
$k$ in $y$, we simulate such inputs with a trivial detection and compute the 
resulting values for the TAUC. We choose an input length of $n=1000$.
When using the step rule, the TAUC is always $0$ as expected, since the only step always
retains its area under the curve of $0$.
But when looking at the obtained TAUC values when using the trapezoidal integration rule,
we can clearly see the TAUC decreasing when $k$ increases. 
\begin{figure}[H]
\centering
\begin{tikzpicture}
        \node[] at (0,0) {\includegraphics[width=0.45\textwidth]{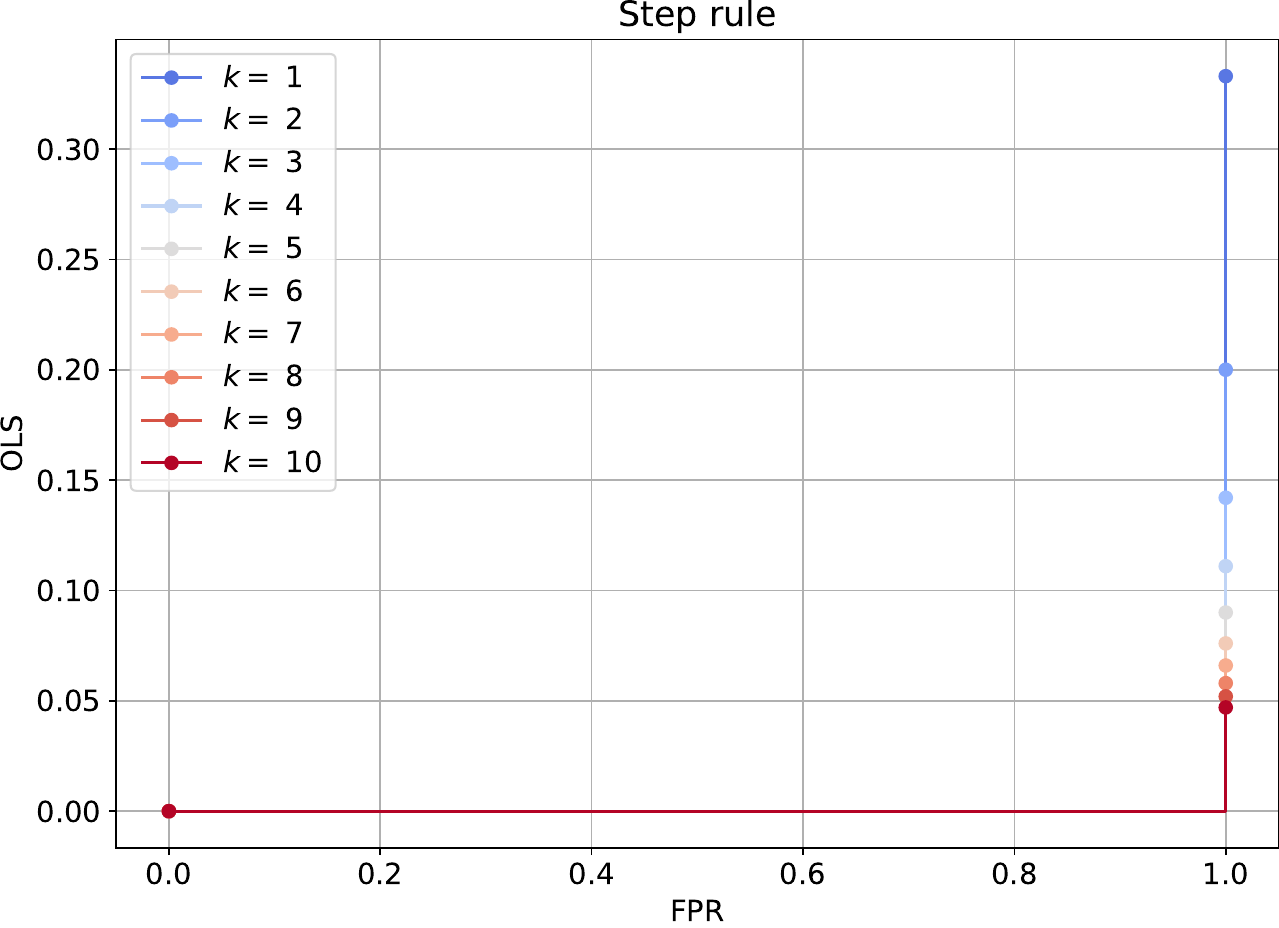}};
        \node[] at (7,0) {\includegraphics[width=0.45\textwidth]{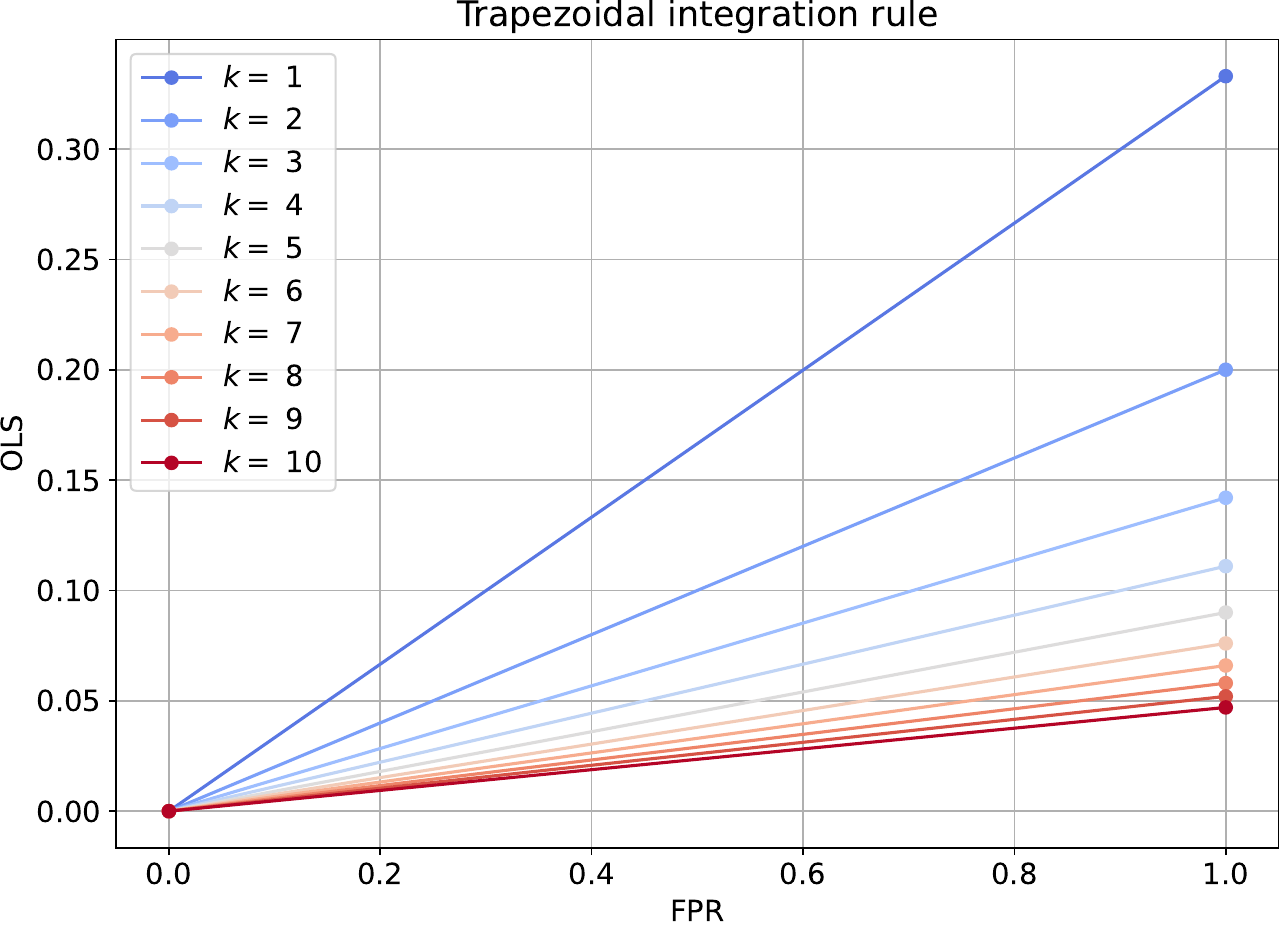}};
        
    \end{tikzpicture}
    \caption{Visualization of the behaviour of the constructed curve for the TAUC on 
    increasing number of segments $k$. 
    The left figure shows that the TAUC for the computation
    with the step rule always remains $0$. The right figure shows that the area under the 
    line decreases with increasing $k$, resulting in a lower TAUC value in case of the
    trapezoidal integration rule.}
\end{figure}
This decreasing behaviour can be approximated by $\frac{1}{2k}$, since the 
TAUC for a trivial detection with $k$ segments in case of the trapezoidal rule can 
be computed with $\frac{P}{2k}$ and $0 < P \leq 1$.
Thus, the limit of the TAUC computed with the trapezoidal integration rule with increasing $k$ follows as:
$$
	\lim_{k \to \infty} \frac{P}{2k} = 0
$$

\begin{figure}[H]
    \centering
    \includegraphics[scale=0.4]{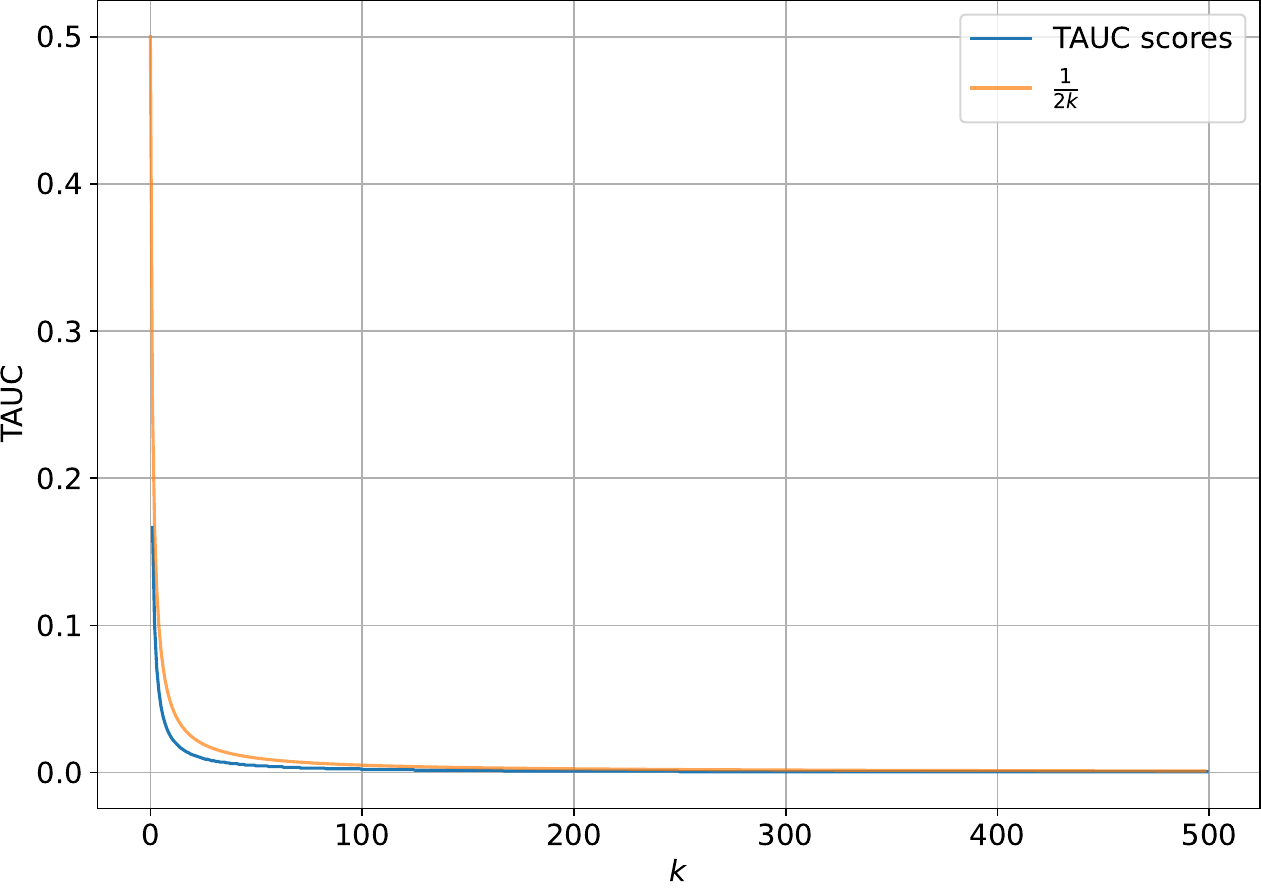}    
    \caption{Visualization of the TAUC with the trapezoidal integration rule, when increasing $k$,
    alongside an approximation $\frac{1}{2k}$. 
    The TAUC gets closer to $0$ with increasing $k$.} 
\end{figure}

\end{document}

%% file: ts_drifts.tex
    \begin{tikzpicture}

        \node[anchor=west,scale=0.8] at (-6,0) {Time series data};
        \node[anchor=west,scale=0.8] at (-6,-2) {Profile data};
        \node[anchor=west,scale=0.8] at (-6,-4) {Process curve data};

        \draw[->,thick] (-3,-5.5) to node[midway, fill=white, align=center, scale=0.8] {Process executions} (3,-5.5);

        \node[] (pd) at (0,-2)  {

            \begin{tikzpicture}
                \node[] at (0,0) {\includegraphics[width=2cm]{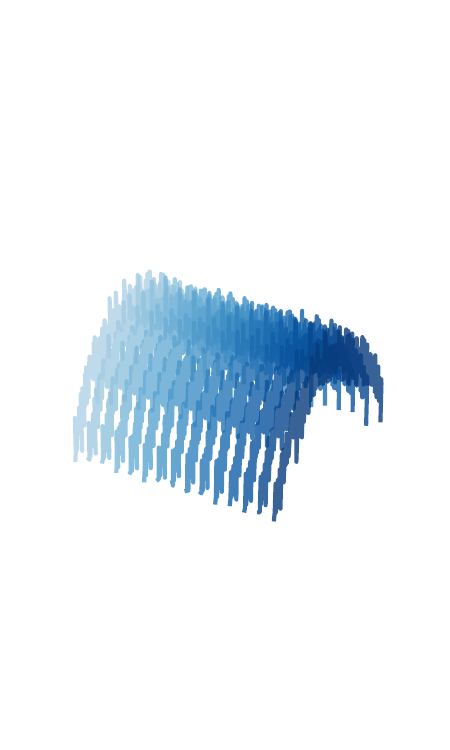}};
                \node[] at (1.5,0) {\includegraphics[width=2cm]{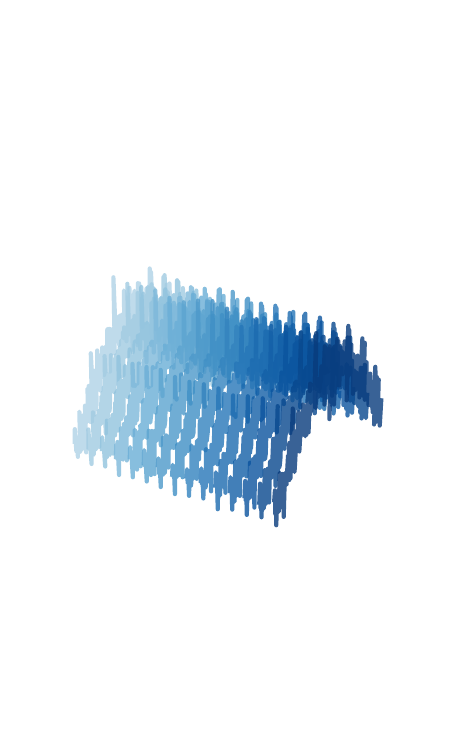}};
                \node[] at (3,0) {\includegraphics[width=2cm]{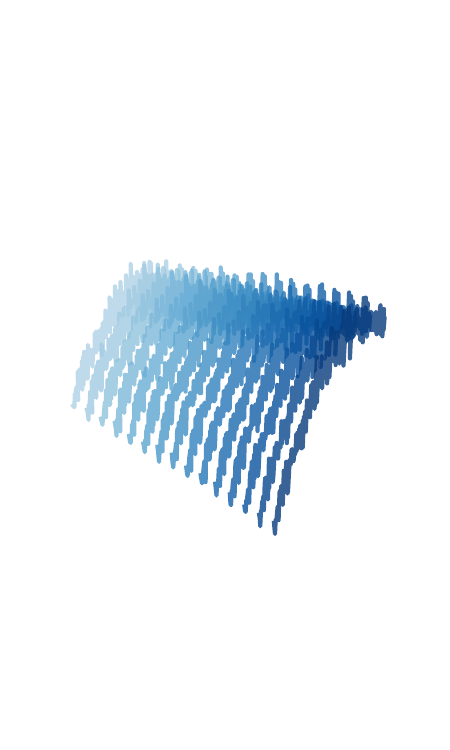}};
                \node[] at (4.5,0) {\includegraphics[width=2cm]{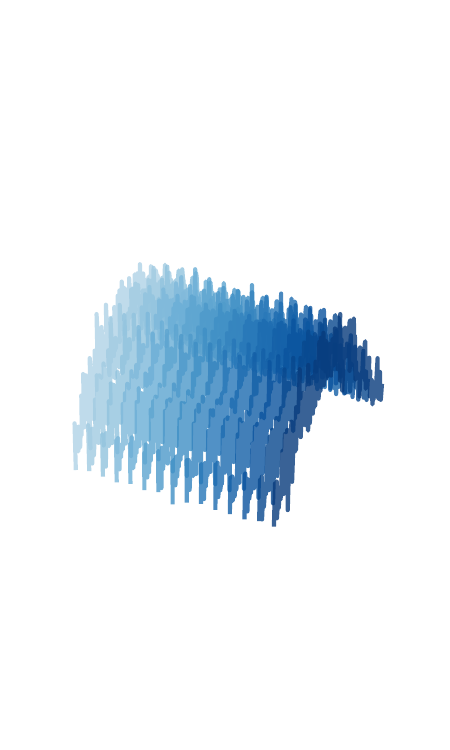}};
                
\draw[->, very thick, color=red] (2.2,-0.3) to node[midway, sloped, scale=0.5, fill=white] {drift} (3.1, -0.81);

            \end{tikzpicture}

        };

        \node[] (ts) at (0,0) {
               \begin{tikzpicture}
                   \node[] at (0,-2) {\includegraphics[width=6cm]{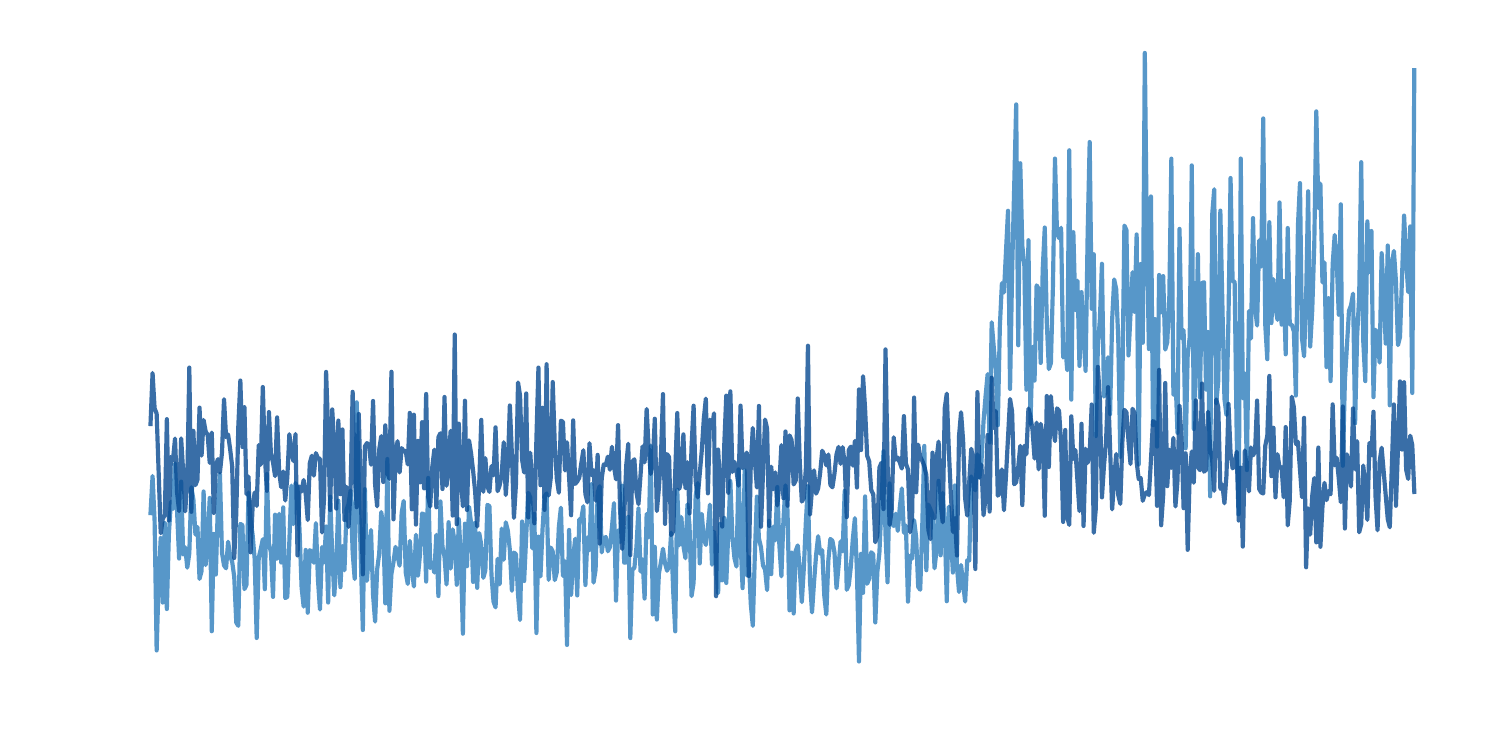}};
                   \draw[->, very thick, color=red] (0.7,-2) to node[midway, sloped, scale=0.5, fill=white] {drift} (1.0, -1);
               \end{tikzpicture}
        };

        \node[] (pc) at (0,-4) {
            \begin{tikzpicture}
                \node[] at (0,0) {\includegraphics[width=2cm]{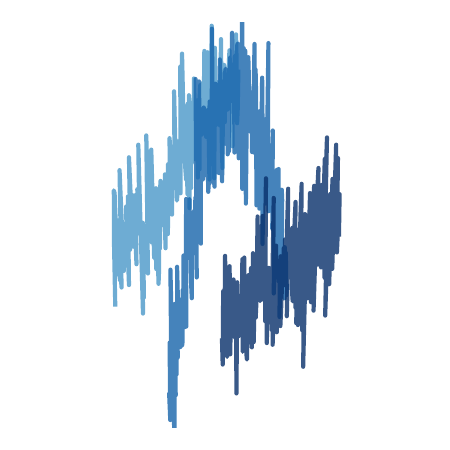}};
                \node[] at (1.5,0) {\includegraphics[width=2cm]{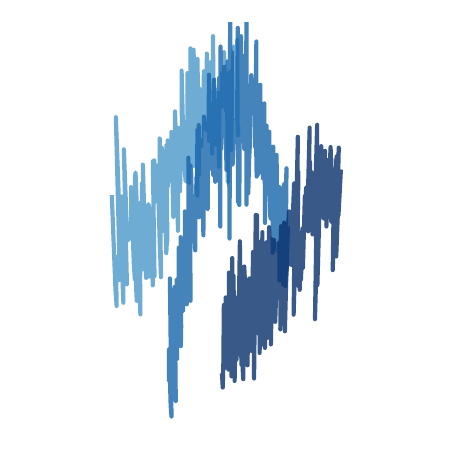}};
                \node[] at (3,0) {\includegraphics[width=2cm]{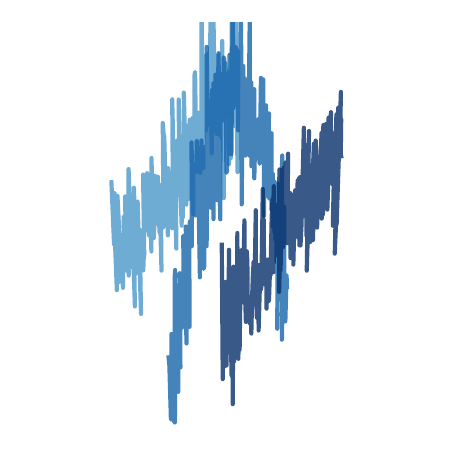}};
                \node[] at (4.5,0) {\includegraphics[width=2cm]{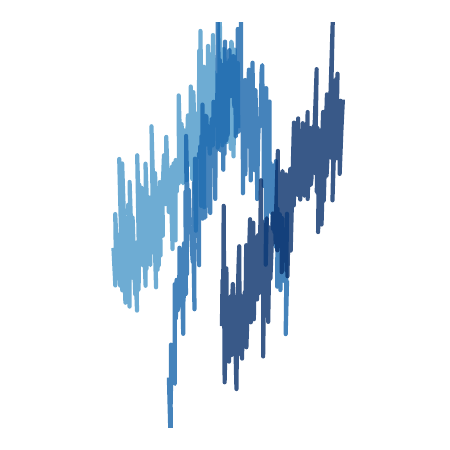}};

\draw[->, very thick, color=red] (3,-1) to node[midway, sloped, scale=0.5, fill=white] {drift} (4.5, -1);

            \end{tikzpicture}

        };
        
    \end{tikzpicture}

%% file: overlap_score_example.tex
\begin{tikzpicture}[scale=2.]
    \draw[->] (0,0) to node[midway, fill=white]{Process executions} (8,0) node[right] {t};
      
      \draw[dashed] (0,0.7) -- (8,0.7) node[right]{$\hat{\cD}(s, \tau)$};
      \draw[dashed] (0,1.4) -- (8,1.4) node[right]{$\cD$};

      \fill[draw=black, fill=green!50!black] (1.5,1.2) rectangle (3.5,1.6);
      \node at (2.5, 1.4) {$\mathcal{D}_1$};
      
      \fill[draw=black, fill=blue!30!white] (1.2,0.5) rectangle (2.5,0.9); 
      \fill[draw=black, fill=blue!30] (3.0,0.5) rectangle (3.8,0.9);
  \node at (1.85, 0.7) {$\mathcal{\hat{D}}_1$};
  \node at (3.4, 0.7) {$\mathcal{\hat{D}}_2$};

      \draw[red, dashed, very thick] (1.2, 0.1) -- (1.2,1.6);
      \draw[red, dashed, very thick] (3.8, 0.1) -- (3.8,1.6);

      \fill[draw=black, fill=green!50!black] (5.0,1.2) rectangle (7.0,1.6);
      \node at (6.0, 1.4) {$\mathcal{D}_2$};
      
      \fill[draw=black, fill=blue!30] (5.5,0.5) rectangle (6.5,0.9); 
      \node at (6.0, 0.7) {$\mathcal{\hat{D}}_3$};

      \draw[red, dashed, very thick] (5.0, 0.1) -- (5.0,1.6);
      \draw[red, dashed, very thick] (7.0, 0.1) -- (7.0,1.6);
    \end{tikzpicture}